\documentclass{article}

\usepackage[preprint]{log_2023-my}

\newif\ifnbld
\newif\ifconf
\newif\iftr
\newif\ifEXT
\EXTfalse

\trtrue
\conffalse

\trfalse
\conftrue

\nbldtrue

\usepackage[numbers,compress,sort]{natbib}

\usepackage[utf8]{inputenc} %
\usepackage[T1]{fontenc}    %
\usepackage{url}            %
\usepackage{booktabs}       %
\usepackage{amsfonts}       %
\usepackage{nicefrac}       %
\usepackage{microtype}      %
\usepackage{xcolor}         %

\usepackage{wrapfig}

\newif\ifsq     %

\newif\ifsqCAP
\newif\ifsqVS
\newif\ifsqEN
\newif\ifsqTIT

\newcommand{\ignore}[1]{}

\sqtrue
\sqCAPfalse
\sqENtrue
\sqVStrue
\sqTITtrue

\sqfalse
\sqCAPfalse
\sqENfalse
\sqVSfalse
\sqTITfalse

\usepackage{etex}
\usepackage{balance}
\usepackage{epstopdf}
\usepackage{placeins}

\usepackage{graphicx}
\usepackage{float}
\usepackage{dblfloatfix}
\usepackage{multirow}
\usepackage{rotating}
\usepackage{makecell}
\usepackage{tabulary}
\usepackage{parcolumns}
\usepackage{tikz}
\usetikzlibrary{tikzmark}

\usepackage{xpatch}
\expandafter\xpatchcmd
\csname pgfk@/tikz/every picture/.@cmd\endcsname
{\thepage}{\arabic{page}}{}{}

\tikzstyle{comment} = [draw, fill=blue!70, text=white, text width=3cm, minimum height=1cm, rounded corners, align=left, font=\scriptsize]
\tikzstyle{background_alg} = [draw, fill=blue!20, opacity=0.4, inner sep=4pt, rounded corners=2pt]

\usetikzlibrary{shapes}
\usetikzlibrary{plotmarks}
\usetikzlibrary{calc, fit}

\usepackage{enumitem}

\usepackage{amsthm}
\usepackage{amsmath,amssymb,amsfonts}
\usepackage{mathtools,mathrsfs}

\usepackage{soul}
\usepackage{fontawesome}
\usepackage{pifont}
\usepackage{textcomp}
\usepackage{booktabs}
\usepackage{url}
\usepackage{pbox}
\usepackage[normalem]{ulem}
\usepackage[10pt]{moresize}

\ifsqCAP
\usepackage[font={normalfont, scriptsize}]{caption}
\usepackage[font={normalfont, scriptsize}]{subcaption}
\else
\usepackage[font={normalfont, scriptsize}]{caption}
\usepackage[font={normalfont, scriptsize}]{subcaption}
\fi

\newcommand{\vspaceSQ}[1]{\ifsqVS\vspace{#1}\fi}
\newcommand{\enlargeSQ}[1]{\ifsqEN\enlargethispage{\baselineskip}\fi}

\ifsqTIT
\usepackage[compact]{titlesec}
\titlespacing*{\section}{0pt}{3pt}{-1pt}
\titlespacing*{\subsection}{0pt}{0pt}{-3pt}
\titlespacing*{\subsubsection}{0pt}{2pt}{1pt}
\fi

\usepackage{xcolor}
\definecolor{darkgrey}{RGB}{70,70,70}
\definecolor{lightgrey}{RGB}{200,200,200}
\definecolor{lyellow}{RGB}{255,255,100}
\definecolor{llyellow}{RGB}{250,250,180}
\definecolor{lgreen}{RGB}{144,238,144}
\definecolor{raphael_comments}{RGB}{13, 145, 24}

\usepackage[customcolors]{hf-tikz}
\hfsetbordercolor{white}
\hfsetfillcolor{vlgray}

\definecolor{vlgray}{rgb}{0.77 0.77 0.77}
\definecolor{ablack}{rgb}{0.2 0.2 0.2}
\definecolor{vllgray}{rgb}{0.9 0.9 0.9}
\definecolor{bblue}{rgb}{0.7 0.7 0.99}

\usepackage{colortbl}

\usepackage{inconsolata}
\usepackage{listings}

\ifsq
\else
\fi

\definecolor{hlL}{rgb}{0.8 0.8 0.99}

\newcounter{highlight}

\newcounter{hlLR}

\newcounter{hlLIR}

\newcounter{hlLIIR}

\newcounter{Ahighlight}

\usepackage{scalerel,stackengine}
\stackMath
\newcommand\rwh[1]{%
\savestack{\tmpbox}{\stretchto{%
  \scaleto{%
        \scalerel*[\widthof{\ensuremath{#1}}]{\kern-.6pt\bigwedge\kern-.6pt}%
                  {\rule[-\textheight/2]{1ex}{\textheight}}%
                              }{\textheight}%
}{0.5ex}}%
\stackon[1pt]{#1}{\tmpbox}%
}

\usepackage[hang,flushmargin]{footmisc}

\renewcommand{\epsilon}{\ensuremath\varepsilon}

\renewcommand{\phi}{\ensuremath{\varphi}}

\newcommand{\R}{\mathbb{R}}
\newcommand{\Z}{\mathbb{Z}}

\newcommand{\bigO}{\mathcal{O}}

\renewcommand{\epsilon}{\ensuremath\varepsilon}

\renewcommand{\phi}{\ensuremath{\varphi}}

\newcommand*\Vector{\mathbf}
\newcommand*\Matrix{\mathbf}
\newcommand*\Mat{\mathbf}

\newcommand{\GradL}[1]{\frac{\partial \mathcal{L}}{\partial #1}}

\newcommand{\src}{\text{src}}
\newcommand{\dst}{\text{dst}}
\newcommand{\fpb}{\frac{\text{FLOP}}{\text{byte}}}

\usepackage{multirow}
\usepackage[backref=page]{hyperref}
\usepackage{cleveref}
\usepackage{todonotes}
\usepackage{makecell}

\usepackage{stmaryrd}

\definecolor{lightgray1}{gray}{0.6}
\definecolor{lightgray2}{gray}{0.8}

\title{Cached Operator Reordering:\\ A Unified View for Fast GNN Training}

\ifnbld

\author[J. Bazinska et al.]
{
Julia Bazinska \\
ETH Zurich \\
Zurich, Switzerland \\
\email{\{firstname\}.\{lastname\}@inf.ethz.ch}\And
Andrei Ivanov \\
ETH Zurich \\
Zurich, Switzerland \\
\email{\{firstname\}.\{lastname\}@inf.ethz.ch}\And
Tal Ben-Nun \\
ETH Zurich \\
Zurich, Switzerland \\
\email{\{firstname\}.\{lastname\}@inf.ethz.ch}\And
Nikoli Dryden \\
ETH Zurich \\
Zurich, Switzerland \\
\email{\{firstname\}.\{lastname\}@inf.ethz.ch}\And
Maciej Besta \\
ETH Zurich \\
Zurich, Switzerland \\
\email{\{firstname\}.\{lastname\}@inf.ethz.ch}\And
Siyuan Shen \\
ETH Zurich \\
Zurich, Switzerland \\
\email{\{firstname\}.\{lastname\}@inf.ethz.ch}\And
Torsten Hoefler \\
ETH Zurich \\
Zurich, Switzerland \\
\email{\{firstname\}.\{lastname\}@inf.ethz.ch}
}
\fi

\begin{document}

\maketitle

\vspaceSQ{-1em}
\begin{abstract}
\vspaceSQ{-1em}
Graph Neural Networks (GNNs) are a powerful tool for handling structured graph data and addressing tasks such as node classification, graph classification, and clustering.
However, the sparse nature of GNN computation poses new challenges for performance optimization compared to traditional deep neural networks. 
We address these challenges by providing a unified view of GNN computation, I/O, and memory.
By analyzing the computational graphs of the Graph Convolutional Network (GCN) and Graph Attention (GAT) layers---two widely used GNN layers---we propose alternative computation strategies.
We present \emph{adaptive operator reordering with caching}, which achieves a speedup of up to 2.43x for GCN compared to the current state-of-the-art. 
Furthermore, an exploration of different caching schemes for GAT yields a speedup of up to 1.94x.
The proposed optimizations save memory, are easily implemented across various hardware platforms, and have the potential to alleviate performance bottlenecks in training large-scale GNN models.
\end{abstract}

\vspaceSQ{-0.5em}
\section{Introduction}
\label{sec:intro}

Neural networks have transformed the way various real-world problems are solved, from computer vision to natural language processing. However, most deep neural network (DNN) approaches do not allow for straightforward processing of structured graph data, such as molecules, social networks or knowledge graphs. 
There exist complex methods allowing for limited processing of such structured data by regular DNNs.

Some examples include bag-of-atoms \cite{bagofatoms}, which represent chemical compounds, or hierarchical processing of 3D point clouds, which allows the inclusion of local neighborhood information in the computation \cite{pointnetpp}. 
Even though these approaches allow processing structured graph data with DNNs, they are lossy representations that do not include the full information on data connections and their properties. 
To overcome these limitations, a new computational paradigm emerged.

Graph Neural Networks (GNNs) are a method of performing neural network computation on graph data. 
They can be used for solving problems such as node classification, graph classification, and clustering~\cite{wu2020comprehensive, zhou2020graph, zhang2020deep, chami2020machine, hamilton2017representation, bronstein2017geometric, besta2021motif, gianinazzi2021learning, scarselli2008graph}. Example applications include recommendations in social networks \cite{social-networks} or document classification in citation networks \cite{gcn}.
Another common use of GNNs is computer vision, where they are used to analyse 3d point clouds \cite{Point-GNN} or to match image key-points \cite{superglue}.

With the processing of larger datasets and models, optimizing the performance of GNNs becomes crucial. 
GNNs differentiate from traditional DNNs in performance due to their predominant use of sparse computation. The input graph adopts a sparse adjacency matrix representation. Operations involving an exchange of information between neighboring nodes rely on this graph structure, and thus, operate on sparse data. As a result, GNN model training and evaluation often face memory-bound challenges, unlike dense models. Consequently, traditional dense-oriented performance optimization methods are often not directly suitable~\cite{besta2022parallel}.

Several factors impact GNN runtime, with memory throughput being only one aspect. Additional considerations include floating point operations executed, potential time-memory trade-offs between saving values for the backward pass, and characteristics of the input graph. Numerous GNN-focused strategies have been developed to address these performance issues, such as operator fusion and reordering \cite{dgnn}, or automatic graph optimization \cite{graphiler}. Nonetheless, these approaches lack a systematic method for analyzing the runtime effects of high-level performance optimization decisions.

In this work, we propose a unified view of the computational graphs, I/O, and memory utilization of the most common GNN layers. This framework enables us to understand the performance implications of various computation schemes and data formats. Guided by this analysis, we propose alternative computation schemes for two widely used GNN layers: the Graph Convolutional Network layer (GCN) \cite{gcn} and the Graph Attention layer (GAT) \cite{gat}.

The main contributions of this work are:
\begin{itemize}
    \item \textbf{Analysis of GNN Performance Factors.} We investigated three aspects that influence GNN performance, which include the optimal selection of sparse data formats, the choice between caching and recomputing intermediate values, and the organization of operations within the computation graph. These considerations shape the performance of GNN implementations depending on the input dataset.

    \item \textbf{Adaptive Computational Scheme.} Drawing insights from theoretical analysis, we leverage the advantages of adaptively selecting caching strategies and operation ordering for the implementation of GCN and GAT. Additionally, our approach integrates the observed impact of selected sparse data formats on a range of empirically evaluated datasets.
    
    \item \textbf{Implementation of Proposed Computation Scheme.} We implemented GAT and GCN within the DaCe~\cite{dace} framework. This implementation yielded a training acceleration of up to 2.43x for GCN and up to 1.94x for GAT compared to the state-of-the-art implementations available in PyTorch Geometric  \cite{pyg} (PyG).
\end{itemize}

\section{Background}
\vspaceSQ{-0.3em}
\label{sec:background}

GNNs are a type of neural networks that are designed to process graph data, such as social networks, citation graphs or spatial structures. In GNN computation, an iterative exchange of information between a node and its neighbors, called \emph{message passing}, takes place. Each node receives messages from its neighbors and computes its new \emph{node feature vector} using an \emph{aggregation function}.
Each message-passing iteration is represented as a layer in the neural network.

Model training is performed similarly to classical DNNs, i.e., the value of the loss function is optimized through the use of a gradient descent method that leverages backpropagation.
Many variations of GNN layers exist. In this work, we narrow our focus to GCNs and GATs.

Typically, graph data consists of \emph{nodes} characterized by some \emph{node features} and are connected by \emph{edges}. Edges can optionally also have descriptors, called \emph{edge weights}, if they are scalars, or \emph{edge features} if they are vector data. In the following sections, the number of vertices is denoted by $n$ and the number of edges by $e$. The typical way of representing input graphs is to provide the so-called \emph{adjacency matrix} $\mathbf{A}=(a_{ij})_{1\leq i,j\leq n}\in \mathbb{R}^{n \times n}$ where $a_{ij} = 1$ if nodes are adjacent and $a_{ij} = 0$ otherwise. 
Furthermore, node features are represented in a \emph{feature matrix} $\mathbf{X} \in \mathbb{R}^{n \times m}$.

\paragraph{GCN} The Graph Convolutional Network layer is a simple GNN operator that is analogous to a convolution operator.
Firstly, the input features for each node are linearly projected into the new feature space, and then, for each node, the features of adjacent nodes are aggregated to create the final output features (\Cref{sec:gcn-fwd,sec:gcn-bwd}). In this work, we consider GCNs with summation as the aggregation operation. 

\paragraph{GAT} The Graph Attention layer is a more complex operator that employs an attention mechanism over the edges connected to each node. After all node features are projected into the new feature space, the \emph{attention weights}, represented as a sparse matrix $\mathcal{A} \in \mathbb{R}^{n \times n}$, are computed for each edge. The attention weight for each edge is computed based on the source and destination node features using another parameterized linear projection (\Cref{sec:gcn-fwd,sec:gcn-bwd}).

\paragraph{SpMM}
\label{sec:spmm}
The \emph{sparse matrix-matrix multiplication} operator is a multiplication of a sparse matrix $\Matrix{A} \in \R^{n \times m} $ and a dense matrix $\Matrix{B} \in \R^{m \times k}$, resulting in a dense matrix $\Matrix{C} \in \R^{n \times k}$, $\Matrix{C} = \Matrix{A} \Matrix{B}$. There exist highly optimized implementations of this operator \cite{cusparse} and it is widely explored in literature \cite{spmm-shi, spmm-vazquez} (\Cref{sec:related-work}). The SpMM operator, as explained in \Cref{sec:background}, is a very common operator in GNNs. It represents the propagation of information between nodes. To understand the computational characteristics of SpMM, we look at \emph{operational intensity}. Operational intensity for SpMMs executed in GNNs on datasets considered in this work is no higher than $1.066 \fpb$  (\Cref{tab:spmm-op-intensity}) which indicates that the computation is memory-bound.

\paragraph{SDDMM}
Another common operator in GNN computation is the \emph{sampled dense-dense matrix multiplication}~\cite{besta2023highgnns}. Given a sparse matrix $\Matrix{A} \in \R^{n \times k} $ and two dense matrices $\Matrix{B} \in \R^{n \times m}$, $\Matrix{C} \in \R^{m \times k}$, we can compute $\Matrix{D} = \Matrix{A} \odot (\Matrix{B} \cdot \Matrix{C})$, where $\odot$ represents the Hadamard product and $\Matrix{D} \in \R^{n \times k}$ is a sparse matrix. 
Similar to SpMM, this subroutine has existing highly optimized implementations \cite{cusparse}. SDDMM is used in the backward pass of GAT. There, we always need to compute it on matrices with shapes $\Mat{A} \in \R^{n \times n}, \Mat{B}\in\R^{n \times f}$ and $\Mat{C} \in \R^{f \times n}$. SDDMM is also memory-bound (\Cref{tab:sddmm-op-intensity}).

\section{Algorithmic View on  Graph Neural Networks}
\label{chap:optimizing}

In this section, we provide a unified view of the computational graph, I/O, and memory for the GCN layer and the GAT layer in order to draw conclusions that would allow us to develop faster GNN implementations. By taking a principled, high-level approach with the awareness of the low-level performance impact, we find new ways to execute GNN layers and save compute.

\subsection{Analysis of the GCN Computational Graph}\label{sec:gcn-analysis}

\begin{figure}
    \centering
    \begin{subfigure}{0.4\textwidth}
        \centering
        \includegraphics[width=0.8\textwidth]{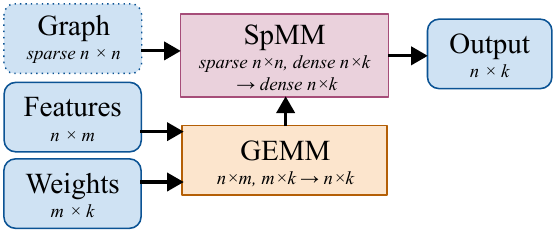}
        \caption{Transform-first GCN forward pass.}
        \label{subfig:gcn-fwd-gemm-first}
    \end{subfigure}
    \begin{subfigure}{0.4\textwidth}
        \centering
        \includegraphics[width=0.8\textwidth]{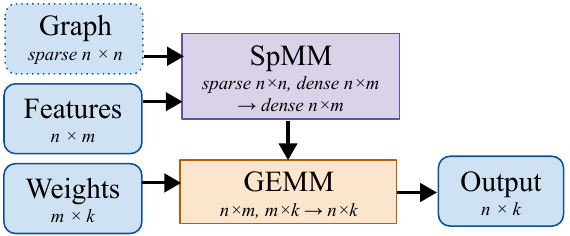}
        \caption{Propagate-first GCN forward pass.}
        \label{subfig:gcn-fwd-spmm-first}
    \end{subfigure}
    \caption{Two schemes of computing the forward pass for GCN. Compute nodes of the same color operate on the same shapes, which are indicated on the node in Einstein summation notation.}
    \label{fig:gcn-fwd-order}
\end{figure}

In order to compute the output of a GCN layer (\Cref{eq:gcn-fwd}), the computation depicted in \Cref{fig:gcn-fwd-order} is necessary. The order of SpMM and GEMM can be altered without impacting the operational outcome. However, it does influence the size of matrices involved in the multiplication. We can either initiate the transformation of features followed by propagation, resulting in the computation of $\mathbf{A} \cdot \left(\mathbf{X} \mathbf{\Theta}\right)$, or opt for the reverse sequence, computing $(\mathbf{A} \cdot \mathbf{X}) \cdot \mathbf{\Theta}$.

The figure reveals that the GEMM is performed on matrices of equal dimensions in both scenarios. The primary difference resides in the SpMM. In the scenario depicted in \Cref{subfig:gcn-fwd-gemm-first} the SpMM operates on matrices sized $n \times n$ and $n \times k$. Conversely, the approach in \Cref{subfig:gcn-fwd-spmm-first} utilizes matrices sized $n \times n$ and $n \times m$. Consequently, the former case entails a computation of $2(qm + nmk)$ FLOPs, while the latter involves $2(qk + nmk)$ FLOPs.

To minimize runtime, the optimal strategy is to initiate feature transformation if $k < m$, and to start with propagation if $k \geq m$. This holds under the assumption that the execution time for an SpMM on matrices of $n \times n$ and $n \times t$ scales linearly with $t$, where $t$ is an integer. This assumption is based on the asymptotic linear scaling of FLOPs and memory transfers in SpMM implementations using common sparse formats (\Cref{tab:spmm-op-intensity}). In reality, the factors influencing runtime are more nuanced and can be dependent on elements such as the exact implementation of the SpMM, the size of the working set, the GPU cache sizes and the sparse matrix structure. 
While we could attempt to model SpMM runtime in a very detailed way, our simpler approach is mostly satisfactory, as can be seen from the results (\Cref{sec:eval}).

\begin{table}[tb]
\caption{Summary of differences between alternative GCN implementations. Entries in the ``Usage condition" column were derived to minimize the total number of operations and memory transfers in the computation. In each pair of alternative schemes, the GEMMs are executed on the same shapes, while the SpMMs are executed on different shapes.}
\centering
\footnotesize
\begin{tabular}{@{}lrrrrr@{}}
\toprule
\textbf{Operation} &
\textbf{Scheme (Figure)} &
\textbf{\makecell[r]{GEMM\\input sizes}} &
\textbf{\makecell[r]{SpMM\\input sizes}} &
\textbf{\makecell[r]{\# Transient\\values}} &
\textbf{\makecell[r]{Usage\\condition}} \\
\midrule
\multirow{2}{*}{GCN forward} &
  \makecell[r]{Transform-first (\ref{subfig:gcn-fwd-gemm-first})} &
  \multirow{2}{*}{\makecell[r]{$nm,mk$}} &
  \begin{tabular}[c]{@{}r@{}}$nn,nk$\end{tabular} &
  $nk$ &
  $k < m $ \\ \cmidrule(lr){2-2} \cmidrule(l){4-6} 
 &
  \begin{tabular}[c]{@{}r@{}}Propagate-first (\ref{subfig:gcn-fwd-spmm-first})\end{tabular} &
   &
  \begin{tabular}[c]{@{}r@{}}$nn, nm$\end{tabular} &
  $nm$ &
  $k \geq m $ \\ \midrule
\multirow{2}{*}{\begin{tabular}[c]{@{}l@{}}GCN backward\\ with feature gradient\end{tabular}} &
  \begin{tabular}[c]{@{}r@{}}Fused-propagate (\ref{subfig:gcn-bwd-reuse})\end{tabular} &
  \multirow{2}{*}{\begin{tabular}[c]{@{}r@{}}$nm, mk$\\ $nm, nk$\end{tabular}} &
  \begin{tabular}[c]{@{}r@{}}$nn, nk$\end{tabular} &
  $nk$ &
  $k < 2m $ \\ \cmidrule(lr){2-2} \cmidrule(l){4-6} 
 &
  \begin{tabular}[c]{@{}r@{}}Split-propagate (\ref{subfig:gcn-bwd-2spmm})\end{tabular} &
   &
  \begin{tabular}[c]{@{}r@{}}$2${}$\times${}$: nn, nm$ \end{tabular} &
  $2nm$ &
  $k \geq 2m $ \\ \midrule
\multirow{2}{*}{\begin{tabular}[c]{@{}l@{}}GCN backward\\ no feature gradient\end{tabular}} &
  \begin{tabular}[c]{@{}r@{}}Fused-propagate (\ref{subfig:gcn-bwd-reuse})\end{tabular} &
  \multirow{2}{*}{\begin{tabular}[c]{@{}r@{}}$nm, mk$\\ $nm, nk$\end{tabular}} &
  \begin{tabular}[c]{@{}r@{}}$nn, nk$\end{tabular} &
  $nk$ &
  $k < m $ \\ \cmidrule(lr){2-2} \cmidrule(l){4-6} 
 &
  \begin{tabular}[c]{@{}r@{}}Split-propagate (\ref{subfig:gcn-bwd-2spmm})\end{tabular} &
   &
  \begin{tabular}[c]{@{}r@{}}$nn, nm$\end{tabular} &
  $nm$ &
  $k \geq m $ \\ \bottomrule
\end{tabular}%
\label{tab:gcn-schemes}
\end{table}

Extending this analysis to the backward pass, we summarize our recommendations on which scheme should be employed in \Cref{tab:gcn-schemes}. Thus, we propose an \emph{adaptive computation scheme} that executes the operations according to the recommendations. In addition to FLOPs, we also consider the amount of memory required by intermediate values in the computation, which is also shown to be smaller when the recommendation is followed. This way, we minimize both the amount of computation needed and the required amount of memory for executing the computation.

\begin{table}
\caption{Comparison of GCN computation schemes with caching. The case presented in \Cref{subfig:gcn-fwd-spmm-first,subfig:gcn-bwd-2spmm} is omitted because it always executes more operations than the scheme using caching. In each pair of alternative schemes, the GEMMs are executed on the same shapes, while the SpMMs are executed on different shapes.}
\centering
\footnotesize
\begin{tabular}{@{}lrrrrrr@{}}
\toprule
\textbf{Operation} &
  \multicolumn{1}{r}{\textbf{\makecell[r]{Forward scheme\\ (Figure)}}} &
  \multicolumn{1}{r}{\textbf{\makecell[r]{Backward scheme\\ (Figure)}}} &
  \multicolumn{1}{r}{\textbf{\makecell[r]{GEMM\\input sizes}}} &
  \multicolumn{1}{r}{\textbf{\makecell[r]{SpMM\\input sizes}}} &
  \textbf{\makecell[r]{\# Transients \\ (fwd; bwd)}} &
  \multicolumn{1}{l}{\textbf{\makecell[r]{Usage\\ condition}}} \\ \midrule
\multirow{2}{*}{\makecell[l]{GCN,\\ with feature \\ gradient}} &
  \begin{tabular}[c]{@{}r@{}}Transform-first (\ref{subfig:gcn-fwd-gemm-first})\end{tabular} &
  \begin{tabular}[c]{@{}r@{}}Fused-propagate (\ref{subfig:gcn-bwd-reuse})\end{tabular} &
  \multirow{2}{*}{\begin{tabular}[c]{@{}r@{}}$nm, mk$\\ $nk, mk$\\ $nm, nk$\end{tabular}} &
  \begin{tabular}[c]{@{}r@{}}$2${}$\times${}$: nn, nk$\end{tabular} &
  \begin{tabular}[c]{@{}l@{}}$nk; nk$\end{tabular} &
  $k < m $ \\ \cmidrule(lr){2-3} \cmidrule(l){5-7} 
 &
  \begin{tabular}[c]{@{}r@{}}Propagate-first \\ with caching (\ref{subfig:gcn-fwd-cache})\end{tabular} &
  \begin{tabular}[c]{@{}r@{}}Split-propagate \\ with caching (\ref{subfig:gcn-bwd-cache})\end{tabular} &
   &
  \begin{tabular}[c]{@{}r@{}}$2${}$\times${}$: nn, nm$\end{tabular} &
  \begin{tabular}[c]{@{}l@{}}$nm; nm$\end{tabular} &
  $k \geq m $ \\ \midrule
\multirow{2}{*}{\begin{tabular}[c]{@{}l@{}}GCN,\\ no feature \\ gradient\end{tabular}} &
  \begin{tabular}[c]{@{}r@{}}Transform-first (\ref{subfig:gcn-fwd-gemm-first})\end{tabular} &
  \begin{tabular}[c]{@{}r@{}}Fused-propagate (\ref{subfig:gcn-bwd-reuse})\end{tabular} &
  \multirow{2}{*}{\begin{tabular}[c]{@{}r@{}}$nm, mk$\\ $nm, nk$\end{tabular}} &
  \begin{tabular}[c]{@{}r@{}}$2${}$\times${}$: nn, nk$\end{tabular} &
  \begin{tabular}[c]{@{}l@{}}$nk; nk$\end{tabular} &
  $k < \frac{1}{2}m $ \\ \cmidrule(lr){2-3} \cmidrule(l){5-7} 
 &
  \begin{tabular}[c]{@{}r@{}}Propagate-first \\ with caching (\ref{subfig:gcn-fwd-cache})\end{tabular} &
  \begin{tabular}[c]{@{}r@{}}Split-propagate \\ with caching (\ref{subfig:gcn-bwd-cache})\end{tabular} &
   &
  \begin{tabular}[c]{@{}r@{}}$nn, nm$\end{tabular} &
  \begin{tabular}[c]{@{}l@{}}$nm; 0$\end{tabular} &
  $k \geq \frac{1}{2}m $ \\ \bottomrule
\end{tabular}%
\label{tab:gcn-cached-schemes}
\end{table}

\paragraph{Caching intermediate values}\label{sec:gcn-caching-analysis}
The SpMM result $\Mat{A} \Mat{X}$ appears both in forward (\Cref{eq:gcn-fwd}) and backward (\Cref{eq:gcn-bwd-weights}) passes. Thus, to avoid recomputing, the result can be cached and reused in the backward pass. Caching the aggregated features $\Mat{A} \Mat{X}$ allows to compute one less SpMM.  
In \Cref{tab:gcn-cached-schemes} we present an analysis of the caching scheme illustrated in \Cref{fig:gcn-scheme-cache}.

In the case described here, caching adds no memory overhead as it avoids storing some intermediates (\Cref{sec:no-mem-overhead}). Moreover, as can be seen in Table \ref{tab:gcn-cached-schemes}, computing GCN with caching requires less memory for transients than the same scheme without caching.

\paragraph{Generalizability of the adaptive scheme} 
The proposed adaptive scheme with caching can be applied to any GNN computation that includes the chained multiplication depicted in \Cref{fig:gcn-fwd-order}. Therefore, our findings could be applicable to other GNN layers employing similar chained multiplication, for example, the Graph Isomorphism Network layer~\cite{gin} or the GraphSAGE operator~\cite{graphsage}. This approach is completely hardware-agnostic, making it suitable for CPU computation, GPU computation, and other accelerators. Although our work operates under the assumption of a summation-based aggregation function for neighboring nodes, it can be extended to accommodate any aggregation function that has the distributive property, e.g. mean.

\subsection{Analysis of the GAT Computation}\label{sec:semibatched}

The computational characteristics of GAT are different from those of GCN (\Cref{sec:gat-fwd}). We cannot use a method analogous to what was shown for GCN in subsection \ref{sec:gcn-analysis} because it requires computing $\mathcal{A} \left( \Matrix{X}\Matrix{\Theta} \right)$, where $\mathcal{A}$ is calculated using the value of $\Matrix{X}\Matrix{\Theta}$. Another difference in relation to GCN is that GAT does not consider edge weights of the input graph. Instead, the attention weights, which are computed based on node features, are used.

Similar to \Cref{sec:gcn-analysis}, we investigated other opportunities for operator reordering in the GAT operator (\Cref{sec:gat-op-reord}). We concluded that the optimal scheme is not dependent on the input and is already in use in existing frameworks.

\paragraph{Multi-head GAT} 
GAT layers are often used with multiple heads. 
From the computational perspective, that results in almost every operator in the computation having an additional dimension. 
From the performance perspective, it can mitigate the issue of uncoalesced memory accesses if the data is arranged correctly: depending on the computation, the head dimension should be either last or penultimate. Usually when operating on the graph structure, the program often needs to access the memory holding values for neighbors of a given node, which are not contiguous. Thus, when a cache line is loaded, only a single value from it is used because others belong to other nodes. Once enough heads in GAT are used, the loaded cache line holds values for the same node but of different heads. All of those values are used, because the same computation has to be executed on all heads.

Moreover, a model with multiple heads implies that we need to execute SpMM and SDDMM in a \emph{semi-batched} fashion, meaning that the sparse structure is shared across the batch. In the case of SpMM, we need to multiply $h$ sparse matrices of shapes $n \times n$ that all share the same sparse structure with a dense matrix of shape $n \times h \times k$, where the middle dimension is the batch dimension.

Similarly, regarding SDDMM, we multiply $h$ sparse matrices sharing the same sparse structure with a result of a multiplication of two dense matrices batched along the middle dimension. 
Having the batch dimension not as the leading dimension, but as the middle dimension potentially allows for more contiguous memory accesses. However, such semi-batched computation is not directly supported by optimized vendor libraries such as CuSPARSE~\cite{cusparse}. 

\paragraph{Caching intermediate values}
We analyze caching opportunities for the forward and backward passes of GAT (\Cref{sec:gat-caching}). Caching more variables enables us to reduce computation time in the backward pass, although it does entail a memory cost. The trade-offs discovered between computational benefits and memory costs are summarized in \Cref{tab:gat-caching}.
We evaluate different caching strategies in subsection \ref{sec:exp-gat-caching}.

\begin{table}[]
\caption{Comparison of time-memory trade-offs in GAT.}
\footnotesize
\centering
\begin{tabular}{@{}lrrr@{}}
\toprule
\textbf{Cached values} &
  \multicolumn{1}{r}{\textbf{Additional memory use}} &
  \multicolumn{1}{r}{\textbf{\begin{tabular}[c]{@{}r@{}}Saved FLOPs\end{tabular}}} &
  \multicolumn{1}{r}{\textbf{\begin{tabular}[c]{@{}r@{}}Saved I/O\end{tabular}}} \\ \midrule
\begin{tabular}[c]{@{}l@{}}Transformed features\end{tabular}                                        & $4nhk$      & $\bigO(nhkm)$      & $\bigO(nm + mhk + nhk)$      \\ \midrule
\begin{tabular}[c]{@{}l@{}}Transformed features, \\ node attention\end{tabular}                   & $4nh(k+2)$  & $\bigO(nhkm)$      & $\bigO(nm + mhk + nhk)$      \\ \midrule
\begin{tabular}[c]{@{}l@{}}Transformed features,\\edge attention weights,\\ edge mask\end{tabular} & $4nhk +5qh$ & $\bigO(nhkm + qh)$ & $\bigO(nm + mhk + nhk + qh)$ \\ \bottomrule
\end{tabular}%
\label{tab:gat-caching}
\end{table}

\vspaceSQ{-1em}
\section{Evaluation}
\label{sec:eval}

The network architecture used for GCN evaluation was a simple GNN network with two GCN layers and a Rectified Linear Unit (ReLU) activation between them. For GAT, the network architecture consisted of two GAT layers with 8 heads each and an Exponential Linear Unit (ELU) activation function between them. The sizes of the internal hidden representations vary between experiments in order to benchmark different model sizes. For computing the parameter gradients, the mean squared error loss function was used for full networks and a simple sum in case of~single layer benchmarking.

\paragraph{Baseline} 
Implementations of GAT and GCN layers from PyTorch Geometric (PyG) were used as baselines. Starting from Pytorch 2.0 and PyTorch Geometric 2.3.0, it is possible to compile the model code to obtain a much faster model. This functionality is supported only for the COO format (\Cref{sec:sparse-formats}). Thus, in this work, we report the numbers for compiled PyTorch models with the COO format and not compiled models with CSR format. Additionally, for GAT, we report the results for the dGNN \cite{dgnn} implementation of the operator, which is also available as part of PyTorch Geometric. The dGNN implementation of GAT requires the graph matrix to be stored in both CSC and CSR, thus storing the graph data twice. The dGNN implementation of GAT consists of hand-written CUDA kernels.

\paragraph{Benchmarking}
All of the following experiments were executed on a machine with Intel(R) 6130 @ 2.10GHz CPU, 1.5 TB RAM, and an NVIDIA Tesla V100 16GB PCIe GPU. Each benchmarked computation was run with 10 warm-up iterations and then executed 100 times in 10 blocks in the case of the forward pass or 20 times in 5 blocks in the case of the backward pass (lower because of long execution times) in order to minimize measurement overheads. All reported results for runtimes are medians of $\frac{1}{100}$ or $\frac{1}{20}$ of block execution times. Standard deviation is also plotted in the figures but is mostly unnoticeable due to low variance in the results. The datasets used in benchmarking are summarized in Table \ref{tab:graph_datasets}.

\begin{table}[]
\caption{Graph datasets used in this work. }
\centering
\footnotesize
\begin{tabular}{@{}lrrrrrr@{}}
\toprule
\textbf{Dataset}                       & \textbf{Nodes} & \textbf{Edges} & \textbf{Features} & \textbf{\% NNZ} & \textbf{Classes} & \textbf{Avg. node degree} \\ \midrule
Cora \cite{dataset-planetoid}     & 2,708   & 10,556    & 1,433 & 0.144\% & 7  & 7.8   \\
Citeseer \cite{dataset-planetoid} & 3,327   & 9,104     & 3,703 & 0.082\% & 6  & 5.47  \\
PubMed \cite{dataset-planetoid}   & 19,717  & 88,648    & 500   & 0.023\% & 3  & 8.99  \\
Flickr \cite{dataset-flickr}      & 89,250  & 899,756   & 500   & 0.011\% & 7  & 5.47  \\
OGB-Arxiv \cite{dataset-ogb}      & 169,343 & 1,166,243 & 128   & 0.004\% & 40 & 13.77 \\
\bottomrule
\end{tabular}%
\label{tab:graph_datasets}
\end{table}

\subsection{Graph Convolutional Network}\label{sec:exp-gcn-schemes}

Our proposed method of choosing the optimal implementation depends on the number of input and output features. Thus, we execute a single GCN layer with varying numbers of output features between 8 and 1024. We use the OGB Arxiv dataset, which has 128 node features, and evaluate layers with varied output sizes. We assess both situations, one where input feature gradients are required and another where they are not necessary. We provide the results for schemes both with and without caching. The dataset is represented in the CSC format.

The threshold for switching between the schemes turned out to be 128 or 256 output features (\Cref{sec:gcn-eval-details}), depending on the computation: whether it is performed without or with feature gradient computation. This threshold matches the suggested usage conditions from \Cref{tab:gcn-schemes,tab:gcn-cached-schemes}. Thus, our adaptive implementation can correctly identify this threshold. As a result, it performs as fast as the fastest scheme.

\begin{figure}
    \centering
    \includegraphics[width=.7\textwidth]{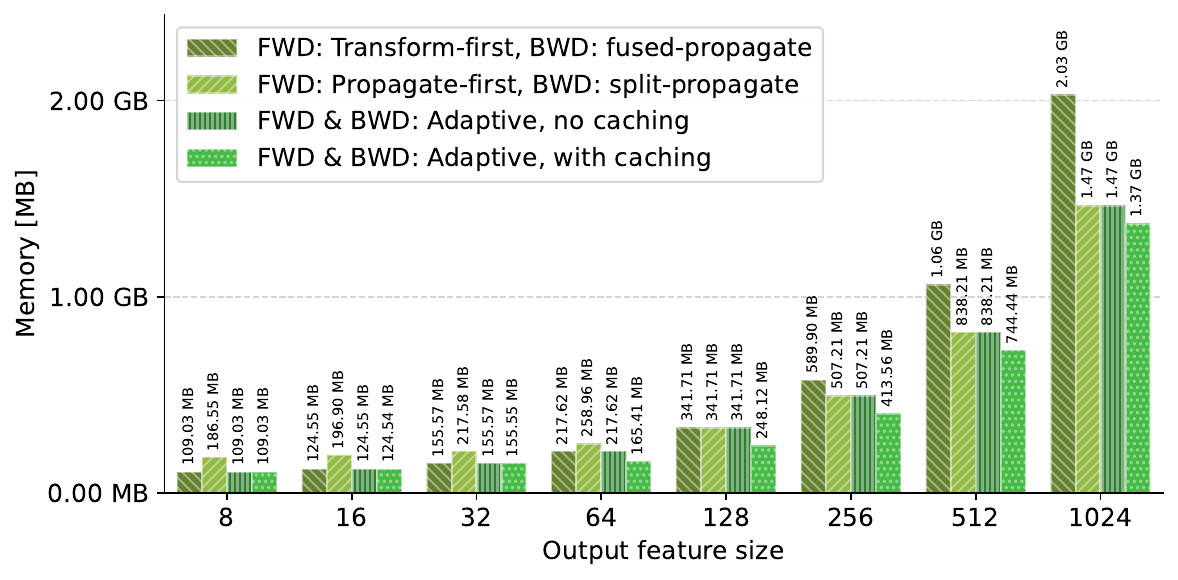}
    \caption{Single GCN layer memory use on the OGB Arxiv dataset.}
    \label{fig:exp-gcn-mem}
\end{figure}

\begin{figure}
\centering
\begin{minipage}{.45\textwidth}
  \centering
  \includegraphics[width=1.0\textwidth]{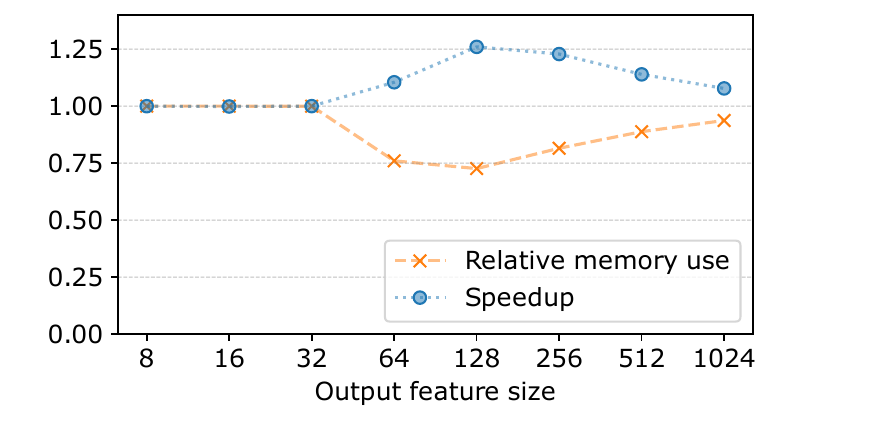}
  \caption{GCN runtime speedup (higher is better) and used memory (lower is better) of the adaptive scheme with caching in comparison to the adaptive scheme without caching.}
  \label{fig:exp-gcn-caching-benefits}
\end{minipage}%
\qquad
\begin{minipage}{.45\textwidth}
  \centering
  \includegraphics[width=1.0\textwidth]{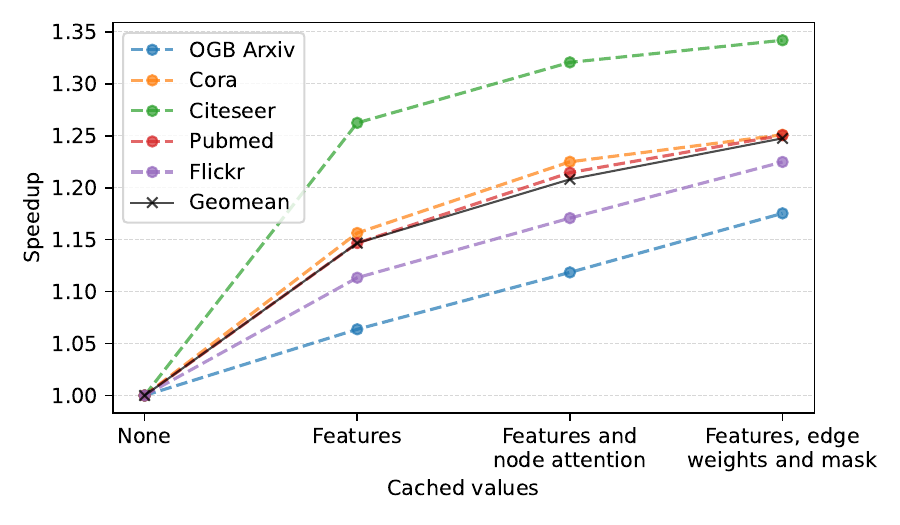}
  \caption{Speedup between different GAT caching schemes for each dataset. Geometric mean is indicated in black.}
  \label{fig:exp-gat-cache-speedup}
\end{minipage}
\end{figure}

\paragraph{Memory usage.} In Section \ref{sec:gcn-caching-analysis}, we mentioned that caching intermediate features does not add extra memory cost. This is backed by Figure \ref{fig:exp-gcn-mem}, which shows memory savings from caching as expected. Caching avoids one SpMM operation, removing the need for an extra array of size $n \times m$. Figure \ref{fig:exp-gcn-caching-benefits} illustrates speedup and memory use from caching versus not caching in the same scheme. For output sizes up to 32, both programs act similarly due to the same transform-first, fused-propagate scheme. With larger output sizes, caching becomes valuable: it exhibits a speedup of up to 25\% while using 25\% less memory. However, for bigger output sizes, the caching advantage lessens as the runtime is dominated by GEMM execution. Thus, adaptive computation benefits certain models, and if the computation scheme allows it, caching intermediate results is always useful.

\paragraph{Evaluation against baselines}

\begin{figure}
    \centering
    \includegraphics[width=.7\textwidth]{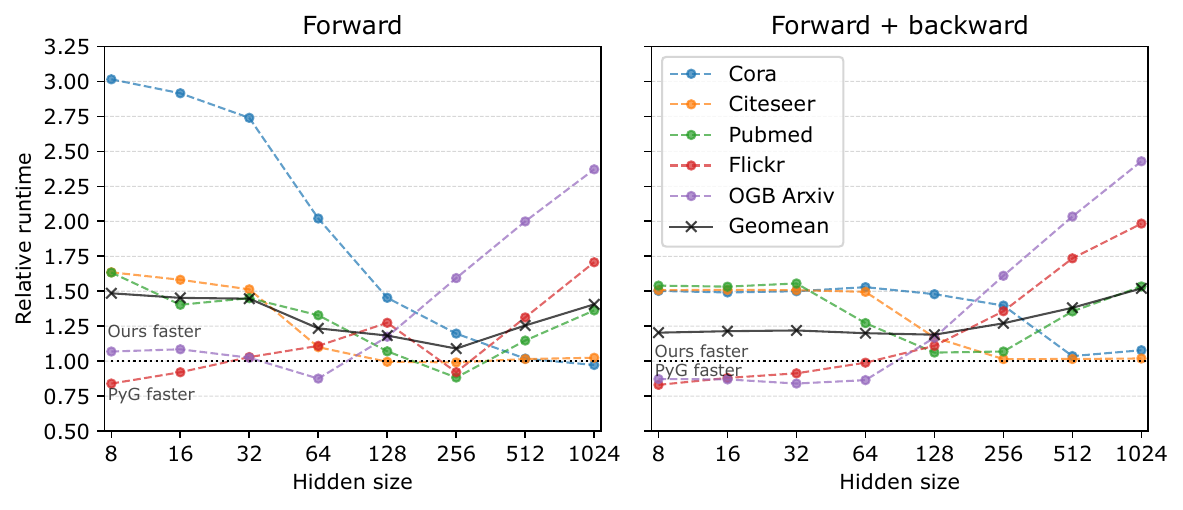}
    \caption{Runtime of our adaptive scheme with caching for GCN relative to the fastest PyTorch Geometric runtime. Geometric mean is indicated in black. }
    \label{fig:exp-gcn-summary}
\end{figure}

After proving our adaptive implementation's optimal computing scheme, we compare it to the baselines on the mentioned two-layer GCN network. We measure the time for a complete forward and backward pass. Runtime results are summarized in \Cref{fig:exp-gcn-summary} (more in \Cref{fig:gcn-main-results}).

In cases where the hidden size is bigger than the input feature size, we benefit greatly from the adaptive scheme. PyTorch Geometric uses only the transform-first scheme for forward and fused-propagate for backward which clearly stands out in the experimental results. At times, our implementation is slower than PyG. This arises when our approach mirrors PyG's computation strategy, specifically the transform-first and fused-propagate methods. The primary distinction between our implementation and PyG pertains to the computation of SpMM and ReLU operations. Our method relies on the CuSPARSE library's optimized subroutine for SpMM, but this prevents us from seamlessly fusing the operator with the subsequent element-wise activation function. In contrast, PyG sacrifices computational flexibility for fine-tuned operator performance. It dynamically generates SpMM code, enabling fusion with the subsequent element-wise operator.

For hidden sizes up to 64, our approach proves faster for smaller datasets like Cora, Citeseer, and Pubmed. Conversely, PyG's approach excels with larger datasets such as Flickr and OGB Arxiv. For increased hidden sizes, we either benefit from an alternative computing scheme, as seen in Arxiv, Flickr, and Pubmed, or achieve performance nearly on par with PyG, exemplified by Citeseer and Cora. This leads us to conclude that CuSPARSE's SpMM likely harnesses the small dataset size to achieve higher performance. However, when data grows too large, CuSPARSE's SpMM performance aligns with PyG's generated SpMM, albeit without the advantage of fused activation function integration.

\begin{figure}
    \centering
    \includegraphics[width=0.7\textwidth]{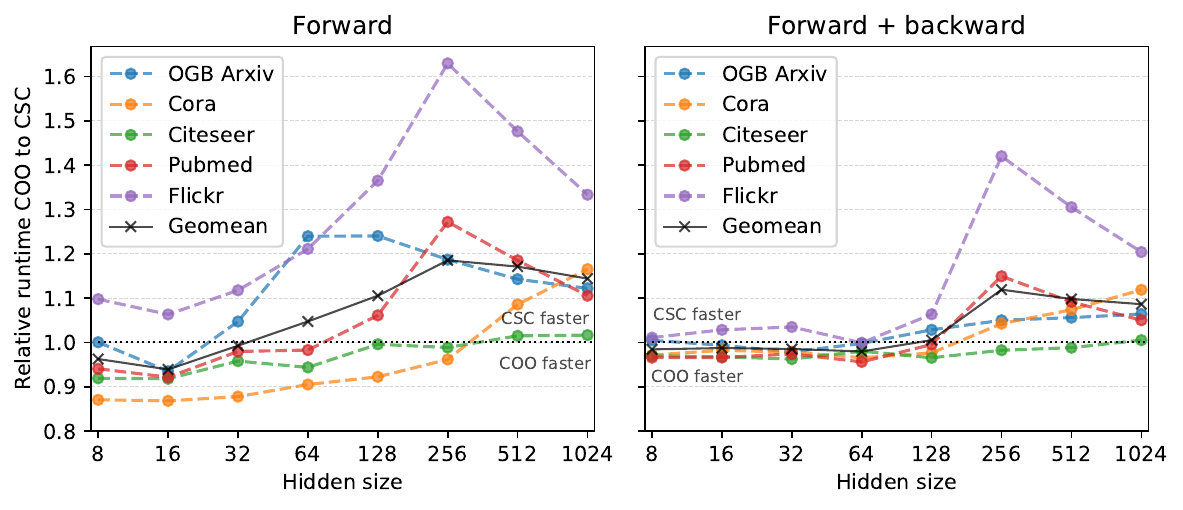}
    \caption{Data format comparison: relative runtime of GCN using the COO format to GCN using the CSC format. Values above $1.0$ indicate that using CSC is faster, below $1.0$ indicate that using COO is faster. Geometric mean across datasets is plotted in black. There is a high variance between datasets but it can be seen that the larger the hidden size, the faster CSC tends to be.}
    \label{fig:exp-gcn-formats}
\end{figure}

The data format choice significantly affects runtime. Comparing COO and CSC formats (see \Cref{fig:exp-gcn-formats}), CSC is up to 1.63 times faster than COO. On average, CSC is 1.14 times faster for forward passes and 1.07 times faster for both passes combined. This impact varies among datasets, with larger hidden sizes favoring CSC due to CuSPARSE's closed-source CSC SpMM subroutines. These utilize a preprocessing step, distributing GPU thread work fairly. As hidden sizes grow, preprocessing takes less time relative to matrix multiplication, reducing overhead for bigger problems. Unlike CSC, COO SpMM variant doesn't need this step, performing better for smaller problems like Cora (hidden size 8). However, for larger graphs, CSC excels by loading less data for sparse representation and benefiting from organized non-zero entries in columns, thus enhancing coalescence.

\subsection{Graph Attention Network}

We benchmark GAT with different caching schemes in order to evaluate their time-memory trade-offs. We run a 2-layered GAT model with 8 heads and varied number of hidden features. We evaluate hidden sizes of 8, 16, 32, 64, 128. A model with 8 heads and hidden size of 128 has a comparable number of parameters as a GCN model with 1024 hidden size. We compare against various implementations available in PyTorch Geometric: compiled COO, CSR, dGNN using CSC and CSR, both compiled and not.

\paragraph{Caching schemes evaluation}\label{sec:exp-gat-caching}
We evaluate four caching schemes as described in Table \ref{tab:gat-caching}: no caching, caching only the transformed features, caching features and node attention, caching features together with final edge weights and LeakyReLU mask. 
Aggregated results for different datasets can be seen in \Cref{fig:exp-gat-cache-speedup} (more in \Cref{fig:exp-gat-mem,fig:exp-gat-caching-results}).

\paragraph{Evaluation against baselines}

\begin{figure}
    \centering
    \includegraphics[width=.7\textwidth]{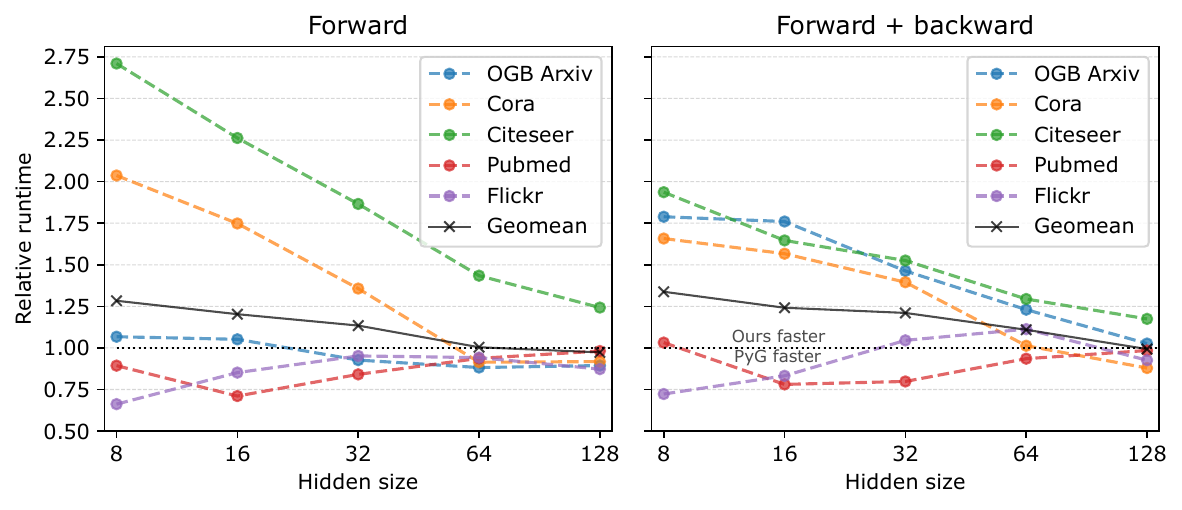}
    \caption{GAT comparison against baselines: the runtime of our implementation relative to the fastest PyTorch Geometric implementation for the given dataset and hidden size. Values above~$1.0$ indicate that ours is faster. The geometric mean of the speedups is indicated in black. }
    \label{fig:exp-gat-summary}
\end{figure}

We choose the fastest caching scheme, which caches the transformed features, edge weights, and the LeakyReLU mask, to benchmark against PyG. We summarize the results in \Cref{fig:exp-gat-summary} (more in \Cref{fig:exp-gat-baselines-results}), where we show the speedup of our implementation against the fastest of PyG implementations for the given dataset and hidden size.

The PyG implementations use different caching strategies. Compiled edge list and CSR both cache as much as possible, i.e., the transformed features, the edge weights, and the edge mask. In the dGNN implementation, only the transformed features and the node attention are cached. Our implementation outperforms the baselines on average. Similar to the GCN case, our implementation is consistently faster than pure PyG implementations on the smallest dataset and small data sizes.

In all instances except one (Flickr, 128 hidden size), our performance surpasses that of the dGNN implementation. We attribute this trend to two key factors. Firstly, we cache full edge weights, whereas dGNN saves only the node attention and recomputes the values. Secondly, the smaller the dataset and the hidden size, the more dGNN is outperformed by our implementation and vanilla PyG. This highlights the suboptimal handling of smaller computations by dGNN, stemming from inadequate utilization of shared memory resources and a lack of adaptability in kernel-blocking strategies for small data. A closer inspection of the source code further substantiates these observations~\cite{dgnn-source}.

However, in some cases our implementation does not outperform PyTorch. That is caused by the fact that our implementation, by committing to highly-optimized subroutines, gives up on possible fusion with sparse operators, which is leveraged by PyTorch, similarly to the case of GCN. Moreover, in multi-head GAT layers, we execute some computation only as batched in the leading dimension, which requires us to execute additional tensor permutations. It would be more efficient to omit the tensor permutations and execute the computation batched in the middle dimension, which would allow for more coalesced memory accesses. However, we take the former approach due to lack of support for the appropriate batching scheme in CuSPARSE. Future work could overcome this difficulty and produce a more optimized GAT implementation.

Furthermore, the experimental results highlight the need for improvements in compilation support within PyTorch Geometric. We encountered difficulties in compiling the GAT model for both Citeseer and OGB Arxiv due to framework bugs in PyTorch. Additionally, for dGNN, we examined results from both compiled and non-compiled models. Surprisingly, we observed that contrary to expectations, the compiled version performs slower than the non-compiled counterpart on the smallest datasets (Cora and Citeseer). This phenomenon is limited to small graphs and hidden sizes. We hypothesize that the PyTorch compiler might generate suboptimal blocking schemes for CUDA kernels in these cases, leading to underutilization of the GPU.

\section{Related Work}
\label{sec:related-work}

\paragraph{Optimizing Sparse Operators} \label{sec:rel-operators}
While our work focuses on high-level GNN computation, the problem of optimizing sparse computation on the operator level has been explored by many works. However, approaches presented in the works below focus solely on low-level optimization of SpMM and thus are not easily generalizable to other computations required by GNNs.

Various approaches aim to optimize sparse matrix-vector multiplication (SpMV) in different contexts. Kreutzer et al. \cite{sell-c-sigma} propose SELL-$C$-$\sigma$, enhancing ELLPACK for efficient SpMV. Anzt et al. \cite{sell-p} present another ELLPACK variant optimized for SpMV. Multiple other works in this direction exist~\cite{besta2017slimsell, besta2017push, solomonik2017scaling, besta2020communication, gianinazzi2018communication}. These solutions offer inspiration rather than direct solutions due to the SpMV absence in GNNs.

Gale et al. \cite{sparse-dl-gale} optimize SpMM and SDDMM for Deep Learning. Their focus is on matrices for pruned dense neural networks, with sparsity of 70\%-90\%. Yet, this is significantly lower than typical matrices representing graphs ($<2\%$ non-zero entries). 

For GNNs, Shi et al. \cite{spmm-shi} propose a SpMM-optimized format using modified COO. Vazquez et al. \cite{spmm-vazquez} design an optimized SpMM kernel using ELLPACK-R. Both address thread balance, memory latency, and uncoalesced reads. GE-SpMM \cite{gespmm} also optimizes SpMM for GNNs, allowing various reduce operators and operating on widely-used CSR format. Recently, Besta et al.~\cite{besta2023highgnns} provided a tensor-based formulation for a broad set of Attentional GNNs.

\paragraph{High-level optimization}
In our work, we utilize the DaCeML~\cite{daceml} and DaCe frameworks to enable high-level optimization beyond the scope of a single operator. There are other works that also adopt such a high-level approach to optimizing the GNN runtime.

Zhang et al. \cite{dgnn} use a layer-level approach similar to ours, wherein they examine computation graphs of selected GNN layers to pinpoint performance issues. Their solution involves reordering operators in the GAT to reduce computation, but they do not address the GCN and lack dynamic computation adjustments based on matrix size.

Methods to alleviate high memory bandwidth usage exist, including \emph{neighbor grouping} by Huang et al. \cite{understanding-gnn-huang}. This approach assigns neighbor node groups to memory-sharing threads, thereby enhancing data reuse. It addresses GNN performance in a distinct manner compared to our work.

GNN-focused compilers also address sparse data optimization. Graphiler \cite{graphiler} automatically enhances GPU-based GNN computation by employing operator reordering optimization, similar to our approach and that of Zhang et al. \cite{dgnn}. In terms of CPU execution, Graphite \cite{graphite} optimizes GNNs by overlapping memory and compute tasks to improve data locality. However, it is restricted to CPUs.

SparseTIR \cite{sparsetir} adopts a data-centered perspective, offering composability in terms of formats and transformations. It tailors hybrid data formats and optimizes computation in alignment with its intermediate representation (IR).

\section{Conclusion}

In this work, we presented a unified view on GNN computational graphs, I/O and memory which allowed us to connect a high-level understanding with the awareness of low-level performance consequences. We used the gained insights to propose GNN optimizations, we implemented them and we showed their benefits in terms of performance. 

We proposed the adaptive computational scheme with caching for optimizing the chained multiplication of $\Matrix{A \cdot X \cdot \Theta}$ where $\Mat{A}$ is a sparse adjacency matrix and $\Matrix{X}$ and $\Matrix{\Theta}$ are dense matrices.
This scheme eliminates redundant computation and reuses cached values to save both compute and memory. We incorporated the scheme in the Graph Convolutional Network layer using the DaCe framework and achieved up to 3.02x speedup (geomean 1.31x) in comparison to the best PyTorch implementation in the forward pass and up to 2.43x speedup (geomean 1.27x) in the backward pass. Furthermore, we evaluated GCN runtime on different sparse data formats and found that the choice of a correct format can result in up to 1.63x speedup.

Moreover, we explored alternative caching schemes for GAT and showed an in-depth analysis of the influence of caching on runtime. Our optimized implementation achieved up to 2.71x speedup (geomean 1.11x) in comparison to the best PyTorch Geometric implementation in the forward pass and up to 1.94x speedup (geomean 1.17x) in the backward pass.

Furthermore, this work provides explicit formulations of the backward passes for both GCN and GAT (\Cref{sec:gcn-gat-implementaitons}). Leveraging their mathematical properties was crucial in this work. Thus, further work could also benefit from their availability.
Another contribution of this work is the extension of DaCeML \cite{daceml} which allows for replacements of PyTorch modules with arbitrary code. This functionality facilitates further work in optimization of machine learning models previously not supported by DaCeML, such as GNN models.

\ifnbld

\section*{Acknowledgements}

We thank Hussein Harake, Colin
McMurtrie, Mark Klein, Angelo Mangili, and the whole CSCS team granting access
to the Ault and Daint machines, and for their excellent technical support. We
thank Timo Schneider for help with computing infrastructure at SPCL.
This project received funding from the European Research Council
\raisebox{-0.25em}{\includegraphics[height=1em]{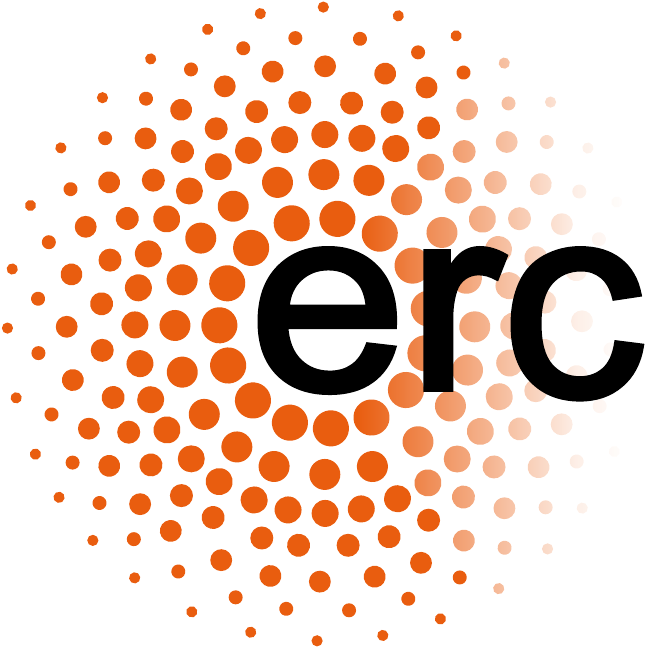}} (Project PSAP,
No.~101002047), and the European High-Performance Computing Joint Undertaking
(JU) under grant agreements No.~955513 (MAELSTROM) and No.~101034126 (EU-Pilot).
This project was supported by the ETH Future Computing Laboratory (EFCL),
financed by a donation from Huawei Technologies.
This project received funding from the European Union's HE research and
innovation programme under the grant agreement No.~101070141 (Project
GLACIATION).

\fi

{
  \footnotesize
\bibliographystyle{ACM-Reference-Format}
\bibliography{references, references_cache}


\newcommand{\SortNoop}[1]{}
\begin{thebibliography}{47}


\ifx \showCODEN    \undefined \def \showCODEN     #1{\unskip}     \fi
\ifx \showDOI      \undefined \def \showDOI       #1{#1}\fi
\ifx \showISBNx    \undefined \def \showISBNx     #1{\unskip}     \fi
\ifx \showISBNxiii \undefined \def \showISBNxiii  #1{\unskip}     \fi
\ifx \showISSN     \undefined \def \showISSN      #1{\unskip}     \fi
\ifx \showLCCN     \undefined \def \showLCCN      #1{\unskip}     \fi
\ifx \shownote     \undefined \def \shownote      #1{#1}          \fi
\ifx \showarticletitle \undefined \def \showarticletitle #1{#1}   \fi
\ifx \showURL      \undefined \def \showURL       {\relax}        \fi
\providecommand\bibfield[2]{#2}
\providecommand\bibinfo[2]{#2}
\providecommand\natexlab[1]{#1}
\providecommand\showeprint[2][]{arXiv:#2}

\bibitem[dgn({[n.\,d.]})]%
        {dgnn-source}
 \bibinfo{year}{[n.\,d.]}\natexlab{}.
\newblock \bibinfo{title}{{dgNN}: High-performance backend for {GNN} layers
  with {Data Flow Graph} level optimization}.
\newblock \bibinfo{howpublished}{\url{https://github.com/dgSPARSE/dgNN/}}.
\newblock
\newblock
\shownote{Accessed: 2023-08-11}.


\bibitem[Antunes et~al\mbox{.}(2022)]%
        {bagofatoms}
\bibfield{author}{\bibinfo{person}{Luis~M Antunes}, \bibinfo{person}{Ricardo
  Grau-Crespo}, {and} \bibinfo{person}{Keith~T Butler}.}
  \bibinfo{year}{2022}\natexlab{}.
\newblock \showarticletitle{Distributed representations of atoms and materials
  for machine learning}.
\newblock \bibinfo{journal}{\emph{npj Computational Materials}}
  \bibinfo{volume}{8}, \bibinfo{number}{1} (\bibinfo{year}{2022}),
  \bibinfo{pages}{1--9}.
\newblock


\bibitem[Anzt et~al\mbox{.}(2014)]%
        {sell-p}
\bibfield{author}{\bibinfo{person}{Hartwig Anzt}, \bibinfo{person}{Stanimire
  Tomov}, {and} \bibinfo{person}{Jack Dongarra}.}
  \bibinfo{year}{2014}\natexlab{}.
\newblock \bibinfo{booktitle}{\emph{Implementing a Sparse Matrix Vector Product
  for the {SELL-C/SELL-C-sigma} formats on {NVIDIA} {GPUs}}}.
\newblock \bibinfo{type}{{T}echnical {R}eport} UT-EECS-14-727.
\newblock


\bibitem[Ben-Nun et~al\mbox{.}(2019)]%
        {dace}
\bibfield{author}{\bibinfo{person}{Tal Ben-Nun}, \bibinfo{person}{Johannes de
  Fine~Licht}, \bibinfo{person}{Alexandros~N Ziogas}, \bibinfo{person}{Timo
  Schneider}, {and} \bibinfo{person}{Torsten Hoefler}.}
  \bibinfo{year}{2019}\natexlab{}.
\newblock \showarticletitle{Stateful dataflow multigraphs: A data-centric model
  for performance portability on heterogeneous architectures}. In
  \bibinfo{booktitle}{\emph{Proceedings of the International Conference for
  High Performance Computing, Networking, Storage and Analysis}}.
  \bibinfo{pages}{1--14}.
\newblock


\bibitem[Besta et~al\mbox{.}(2020)]%
        {besta2020communication}
\bibfield{author}{\bibinfo{person}{Maciej Besta} {et~al\mbox{.}}}
  \bibinfo{year}{2020}\natexlab{}.
\newblock \showarticletitle{Communication-efficient jaccard similarity for
  high-performance distributed genome comparisons}. In
  \bibinfo{booktitle}{\emph{IEEE IPDPS}}. IEEE, \bibinfo{pages}{1122--1132}.
\newblock


\bibitem[Besta et~al\mbox{.}(2023)]%
        {besta2023highgnns}
\bibfield{author}{\bibinfo{person}{Maciej Besta} {et~al\mbox{.}}}
  \bibinfo{year}{2023}\natexlab{}.
\newblock \showarticletitle{High-Performance and Programmable Attentional Graph
  Neural Networks with Global Tensor Formulations}. In
  \bibinfo{booktitle}{\emph{ACM/IEEE Supercomputing}}.
\newblock


\bibitem[Besta et~al\mbox{.}(2022)]%
        {besta2021motif}
\bibfield{author}{\bibinfo{person}{Maciej Besta}, \bibinfo{person}{Raphael
  Grob}, \bibinfo{person}{Cesare Miglioli}, \bibinfo{person}{Nicola Bernold},
  \bibinfo{person}{Grzegorz Kwasniewski}, \bibinfo{person}{Gabriel Gjini},
  \bibinfo{person}{Raghavendra Kanakagiri}, \bibinfo{person}{Saleh Ashkboos},
  \bibinfo{person}{Lukas Gianinazzi}, \bibinfo{person}{Nikoli Dryden},
  {et~al\mbox{.}}} \bibinfo{year}{2022}\natexlab{}.
\newblock \showarticletitle{Motif Prediction with Graph Neural Networks}, In
  \bibinfo{booktitle}{ACM KDD}.
\newblock \bibinfo{journal}{\emph{arXiv preprint arXiv:2106.00761}}.
\newblock


\bibitem[Besta and Hoefler(2022)]%
        {besta2022parallel}
\bibfield{author}{\bibinfo{person}{Maciej Besta} {and} \bibinfo{person}{Torsten
  Hoefler}.} \bibinfo{year}{2022}\natexlab{}.
\newblock \showarticletitle{Parallel and Distributed Graph Neural Networks: An
  In-Depth Concurrency Analysis}.
\newblock \bibinfo{journal}{\emph{arXiv preprint arXiv:2205.09702}}
  (\bibinfo{year}{2022}).
\newblock


\bibitem[Besta et~al\mbox{.}(2017a)]%
        {besta2017slimsell}
\bibfield{author}{\bibinfo{person}{Maciej Besta}, \bibinfo{person}{Florian
  Marending}, \bibinfo{person}{Edgar Solomonik}, {and} \bibinfo{person}{Torsten
  Hoefler}.} \bibinfo{year}{2017}\natexlab{a}.
\newblock \showarticletitle{Slimsell: A vectorizable graph representation for
  breadth-first search}. In \bibinfo{booktitle}{\emph{IEEE IPDPS}}. IEEE,
  \bibinfo{pages}{32--41}.
\newblock


\bibitem[Besta et~al\mbox{.}(2017b)]%
        {besta2017push}
\bibfield{author}{\bibinfo{person}{Maciej Besta}, \bibinfo{person}{Micha{\l}
  Podstawski}, \bibinfo{person}{Linus Groner}, \bibinfo{person}{Edgar
  Solomonik}, {and} \bibinfo{person}{Torsten Hoefler}.}
  \bibinfo{year}{2017}\natexlab{b}.
\newblock \showarticletitle{To push or to pull: On reducing communication and
  synchronization in graph computations}. In \bibinfo{booktitle}{\emph{ACM
  HPDC}}.
\newblock


\bibitem[Bronstein et~al\mbox{.}(2017)]%
        {bronstein2017geometric}
\bibfield{author}{\bibinfo{person}{Michael~M Bronstein}, \bibinfo{person}{Joan
  Bruna}, \bibinfo{person}{Yann LeCun}, \bibinfo{person}{Arthur Szlam}, {and}
  \bibinfo{person}{Pierre Vandergheynst}.} \bibinfo{year}{2017}\natexlab{}.
\newblock \showarticletitle{Geometric deep learning: going beyond euclidean
  data}.
\newblock \bibinfo{journal}{\emph{IEEE Signal Processing Magazine}}
  \bibinfo{volume}{34}, \bibinfo{number}{4} (\bibinfo{year}{2017}),
  \bibinfo{pages}{18--42}.
\newblock


\bibitem[Chami et~al\mbox{.}(2020)]%
        {chami2020machine}
\bibfield{author}{\bibinfo{person}{Ines Chami}, \bibinfo{person}{Sami
  {Abu-El-Haija}}, \bibinfo{person}{Bryan Perozzi},
  \bibinfo{person}{Christopher R{\'e}}, {and} \bibinfo{person}{Kevin Murphy}.}
  \bibinfo{year}{2020}\natexlab{}.
\newblock \showarticletitle{Machine learning on graphs: A model and
  comprehensive taxonomy}.
\newblock \bibinfo{journal}{\emph{arXiv preprint arXiv:2005.03675}}
  (\bibinfo{year}{2020}).
\newblock


\bibitem[Fan et~al\mbox{.}(2019)]%
        {social-networks}
\bibfield{author}{\bibinfo{person}{Wenqi Fan}, \bibinfo{person}{Yao Ma},
  \bibinfo{person}{Qing Li}, \bibinfo{person}{Yuan He}, \bibinfo{person}{Eric
  Zhao}, \bibinfo{person}{Jiliang Tang}, {and} \bibinfo{person}{Dawei Yin}.}
  \bibinfo{year}{2019}\natexlab{}.
\newblock \bibinfo{title}{Graph Neural Networks for Social Recommendation}.
\newblock
\newblock
\urldef\tempurl%
\url{https://doi.org/10.48550/ARXIV.1902.07243}
\showDOI{\tempurl}


\bibitem[Fey and Lenssen(2019)]%
        {pyg}
\bibfield{author}{\bibinfo{person}{Matthias Fey} {and} \bibinfo{person}{Jan~E.
  Lenssen}.} \bibinfo{year}{2019}\natexlab{}.
\newblock \showarticletitle{Fast Graph Representation Learning with {PyTorch
  Geometric}}. In \bibinfo{booktitle}{\emph{ICLR Workshop on Representation
  Learning on Graphs and Manifolds}}.
\newblock


\bibitem[Gale et~al\mbox{.}(2020)]%
        {sparse-dl-gale}
\bibfield{author}{\bibinfo{person}{Trevor Gale}, \bibinfo{person}{Matei
  Zaharia}, \bibinfo{person}{Cliff Young}, {and} \bibinfo{person}{Erich
  Elsen}.} \bibinfo{year}{2020}\natexlab{}.
\newblock \bibinfo{title}{Sparse GPU Kernels for Deep Learning}.
\newblock
\newblock
\showeprint[arxiv]{2006.10901}~[cs.LG]


\bibitem[Gianinazzi et~al\mbox{.}(2021)]%
        {gianinazzi2021learning}
\bibfield{author}{\bibinfo{person}{Lukas Gianinazzi},
  \bibinfo{person}{Maximilian Fries}, \bibinfo{person}{Nikoli Dryden},
  \bibinfo{person}{Tal Ben-Nun}, {and} \bibinfo{person}{Torsten Hoefler}.}
  \bibinfo{year}{2021}\natexlab{}.
\newblock \showarticletitle{Learning Combinatorial Node Labeling Algorithms}.
\newblock \bibinfo{journal}{\emph{arXiv preprint arXiv:2106.03594}}
  (\bibinfo{year}{2021}).
\newblock


\bibitem[Gianinazzi et~al\mbox{.}(2018)]%
        {gianinazzi2018communication}
\bibfield{author}{\bibinfo{person}{Lukas Gianinazzi}, \bibinfo{person}{Pavel
  Kalvoda}, \bibinfo{person}{Alessandro De~Palma}, \bibinfo{person}{Maciej
  Besta}, {and} \bibinfo{person}{Torsten Hoefler}.}
  \bibinfo{year}{2018}\natexlab{}.
\newblock \showarticletitle{Communication-avoiding parallel minimum cuts and
  connected components}. In \bibinfo{booktitle}{\emph{ACM SIGPLAN Notices}}.
\newblock


\bibitem[Gong et~al\mbox{.}(2022)]%
        {graphite}
\bibfield{author}{\bibinfo{person}{Zhangxiaowen Gong},
  \bibinfo{person}{Houxiang Ji}, \bibinfo{person}{Yao Yao},
  \bibinfo{person}{Christopher~W. Fletcher}, \bibinfo{person}{Christopher~J.
  Hughes}, {and} \bibinfo{person}{Josep Torrellas}.}
  \bibinfo{year}{2022}\natexlab{}.
\newblock \showarticletitle{Graphite: Optimizing Graph Neural Networks on
  {CPUs} through Cooperative Software-Hardware Techniques}. In
  \bibinfo{booktitle}{\emph{Proceedings of the 49th Annual International
  Symposium on Computer Architecture}} (New York, New York)
  \emph{(\bibinfo{series}{ISCA '22})}. \bibinfo{publisher}{Association for
  Computing Machinery}, \bibinfo{address}{New York, NY, USA},
  \bibinfo{pages}{916–931}.
\newblock
\showISBNx{9781450386104}
\urldef\tempurl%
\url{https://doi.org/10.1145/3470496.3527403}
\showDOI{\tempurl}


\bibitem[Hamilton et~al\mbox{.}(2017)]%
        {hamilton2017representation}
\bibfield{author}{\bibinfo{person}{William~L Hamilton} {et~al\mbox{.}}}
  \bibinfo{year}{2017}\natexlab{}.
\newblock \showarticletitle{Representation learning on graphs: Methods and
  applications}.
\newblock \bibinfo{journal}{\emph{arXiv preprint arXiv:1709.05584}}
  (\bibinfo{year}{2017}).
\newblock


\bibitem[Hamilton et~al\mbox{.}(2018)]%
        {graphsage}
\bibfield{author}{\bibinfo{person}{William~L. Hamilton}, \bibinfo{person}{Rex
  Ying}, {and} \bibinfo{person}{Jure Leskovec}.}
  \bibinfo{year}{2018}\natexlab{}.
\newblock \bibinfo{title}{Inductive Representation Learning on Large Graphs}.
\newblock
\newblock
\showeprint[arxiv]{1706.02216}~[cs.SI]


\bibitem[Hu et~al\mbox{.}(2020)]%
        {dataset-ogb}
\bibfield{author}{\bibinfo{person}{Weihua Hu}, \bibinfo{person}{Matthias Fey},
  \bibinfo{person}{Marinka Zitnik}, \bibinfo{person}{Yuxiao Dong},
  \bibinfo{person}{Hongyu Ren}, \bibinfo{person}{Bowen Liu},
  \bibinfo{person}{Michele Catasta}, \bibinfo{person}{Jure Leskovec},
  \bibinfo{person}{Regina Barzilay}, \bibinfo{person}{Peter Battaglia},
  \bibinfo{person}{Yoshua Bengio}, \bibinfo{person}{Michael Bronstein},
  \bibinfo{person}{Stephan Günnemann}, \bibinfo{person}{Will Hamilton},
  \bibinfo{person}{Tommi Jaakkola}, \bibinfo{person}{Stefanie Jegelka},
  \bibinfo{person}{Maximilian Nickel}, \bibinfo{person}{Chris Re},
  \bibinfo{person}{Le Song}, \bibinfo{person}{Jian Tang}, \bibinfo{person}{Max
  Welling}, {and} \bibinfo{person}{Rich Zemel}.}
  \bibinfo{year}{2020}\natexlab{}.
\newblock \showarticletitle{{Open Graph Benchmark}: Datasets for Machine
  Learning on Graphs}.
\newblock \bibinfo{journal}{\emph{Advances in Neural Information Processing
  Systems}}  \bibinfo{volume}{2020-December} (\bibinfo{date}{5}
  \bibinfo{year}{2020}).
\newblock
\showISSN{10495258}
\urldef\tempurl%
\url{https://arxiv.org/abs/2005.00687v7}
\showURL{%
\tempurl}


\bibitem[Huang et~al\mbox{.}(2020)]%
        {gespmm}
\bibfield{author}{\bibinfo{person}{Guyue Huang}, \bibinfo{person}{Guohao Dai},
  \bibinfo{person}{Yu Wang}, {and} \bibinfo{person}{Huazhong Yang}.}
  \bibinfo{year}{2020}\natexlab{}.
\newblock \bibinfo{title}{{GE-SpMM}: General-purpose Sparse Matrix-Matrix
  Multiplication on {GPUs} for Graph Neural Networks}.
\newblock
\newblock
\showeprint[arxiv]{2007.03179}~[cs.DC]


\bibitem[Huang et~al\mbox{.}(2021)]%
        {understanding-gnn-huang}
\bibfield{author}{\bibinfo{person}{Kezhao Huang}, \bibinfo{person}{Jidong
  Zhai}, \bibinfo{person}{Zhen Zheng}, \bibinfo{person}{Youngmin Yi}, {and}
  \bibinfo{person}{Xipeng Shen}.} \bibinfo{year}{2021}\natexlab{}.
\newblock \showarticletitle{Understanding and Bridging the Gaps in Current
  {GNN} Performance Optimizations}. In \bibinfo{booktitle}{\emph{Proceedings of
  the 26th ACM SIGPLAN Symposium on Principles and Practice of Parallel
  Programming}} (Virtual Event, Republic of Korea)
  \emph{(\bibinfo{series}{PPoPP '21})}. \bibinfo{publisher}{Association for
  Computing Machinery}, \bibinfo{address}{New York, NY, USA},
  \bibinfo{pages}{119–132}.
\newblock
\showISBNx{9781450382946}
\urldef\tempurl%
\url{https://doi.org/10.1145/3437801.3441585}
\showDOI{\tempurl}


\bibitem[Kiefer and Wolfowitz(1952)]%
        {sgd}
\bibfield{author}{\bibinfo{person}{J. Kiefer} {and} \bibinfo{person}{J.
  Wolfowitz}.} \bibinfo{year}{1952}\natexlab{}.
\newblock \showarticletitle{{Stochastic Estimation of the Maximum of a
  Regression Function}}.
\newblock \bibinfo{journal}{\emph{The Annals of Mathematical Statistics}}
  \bibinfo{volume}{23}, \bibinfo{number}{3} (\bibinfo{year}{1952}),
  \bibinfo{pages}{462 -- 466}.
\newblock
\urldef\tempurl%
\url{https://doi.org/10.1214/aoms/1177729392}
\showDOI{\tempurl}


\bibitem[Kingma and Ba(2017)]%
        {adam}
\bibfield{author}{\bibinfo{person}{Diederik~P. Kingma} {and}
  \bibinfo{person}{Jimmy Ba}.} \bibinfo{year}{2017}\natexlab{}.
\newblock \bibinfo{title}{Adam: A Method for Stochastic Optimization}.
\newblock
\newblock
\showeprint[arxiv]{1412.6980}~[cs.LG]


\bibitem[Kipf and Welling(2016)]%
        {gcn}
\bibfield{author}{\bibinfo{person}{Thomas~N Kipf} {and} \bibinfo{person}{Max
  Welling}.} \bibinfo{year}{2016}\natexlab{}.
\newblock \showarticletitle{Semi-supervised classification with graph
  convolutional networks}.
\newblock \bibinfo{journal}{\emph{arXiv preprint arXiv:1609.02907}}
  (\bibinfo{year}{2016}).
\newblock


\bibitem[Kreutzer et~al\mbox{.}(2014)]%
        {sell-c-sigma}
\bibfield{author}{\bibinfo{person}{Moritz Kreutzer}, \bibinfo{person}{Georg
  Hager}, \bibinfo{person}{Gerhard Wellein}, \bibinfo{person}{Holger Fehske},
  {and} \bibinfo{person}{Alan~R. Bishop}.} \bibinfo{year}{2014}\natexlab{}.
\newblock \showarticletitle{A Unified Sparse Matrix Data Format for Efficient
  General Sparse Matrix-Vector Multiplication on Modern Processors with Wide
  {SIMD} Units}.
\newblock \bibinfo{journal}{\emph{{SIAM} Journal on Scientific Computing}}
  \bibinfo{volume}{36}, \bibinfo{number}{5} (\bibinfo{date}{jan}
  \bibinfo{year}{2014}), \bibinfo{pages}{C401--C423}.
\newblock
\urldef\tempurl%
\url{https://doi.org/10.1137/130930352}
\showDOI{\tempurl}


\bibitem[Maas(2013)]%
        {leakyrelu}
\bibfield{author}{\bibinfo{person}{Andrew~L. Maas}.}
  \bibinfo{year}{2013}\natexlab{}.
\newblock \showarticletitle{Rectifier Nonlinearities Improve Neural Network
  Acoustic Models}.
\newblock
\urldef\tempurl%
\url{https://api.semanticscholar.org/CorpusID:16489696}
\showURL{%
\tempurl}


\bibitem[Naumov et~al\mbox{.}(2010)]%
        {cusparse}
\bibfield{author}{\bibinfo{person}{Maxim Naumov}, \bibinfo{person}{L Chien},
  \bibinfo{person}{Philippe Vandermersch}, {and} \bibinfo{person}{Ujval
  Kapasi}.} \bibinfo{year}{2010}\natexlab{}.
\newblock \showarticletitle{{CUSPARSE} library: A set of basic linear algebra
  subroutines for sparse matrice}. In \bibinfo{booktitle}{\emph{GPU Technology
  Conference}}.
\newblock


\bibitem[Qi et~al\mbox{.}(2017)]%
        {pointnetpp}
\bibfield{author}{\bibinfo{person}{Charles~R. Qi}, \bibinfo{person}{Li Yi},
  \bibinfo{person}{Hao Su}, {and} \bibinfo{person}{Leonidas~J. Guibas}.}
  \bibinfo{year}{2017}\natexlab{}.
\newblock \bibinfo{title}{{PointNet++}: Deep Hierarchical Feature Learning on
  Point Sets in a Metric Space}.
\newblock
\newblock
\urldef\tempurl%
\url{https://doi.org/10.48550/ARXIV.1706.02413}
\showDOI{\tempurl}


\bibitem[Rausch et~al\mbox{.}(2022)]%
        {daceml}
\bibfield{author}{\bibinfo{person}{Oliver Rausch}, \bibinfo{person}{Tal
  Ben-Nun}, \bibinfo{person}{Nikoli Dryden}, \bibinfo{person}{Andrei Ivanov},
  \bibinfo{person}{Shigang Li}, {and} \bibinfo{person}{Torsten Hoefler}.}
  \bibinfo{year}{2022}\natexlab{}.
\newblock \showarticletitle{{DaCeML}: A Data-Centric Optimization Framework for
  Machine Learning}. In \bibinfo{booktitle}{\emph{Proceedings of the 36th ACM
  International Conference on Supercomputing}} \emph{(\bibinfo{series}{ICS
  '22})}.
\newblock


\bibitem[Sarlin et~al\mbox{.}(2019)]%
        {superglue}
\bibfield{author}{\bibinfo{person}{Paul{-}Edouard Sarlin},
  \bibinfo{person}{Daniel DeTone}, \bibinfo{person}{Tomasz Malisiewicz}, {and}
  \bibinfo{person}{Andrew Rabinovich}.} \bibinfo{year}{2019}\natexlab{}.
\newblock \showarticletitle{{SuperGlue}: Learning Feature Matching with Graph
  Neural Networks}.
\newblock \bibinfo{journal}{\emph{CoRR}}  \bibinfo{volume}{abs/1911.11763}
  (\bibinfo{year}{2019}).
\newblock
\showeprint[arXiv]{1911.11763}
\urldef\tempurl%
\url{http://arxiv.org/abs/1911.11763}
\showURL{%
\tempurl}


\bibitem[Scarselli et~al\mbox{.}(2008)]%
        {scarselli2008graph}
\bibfield{author}{\bibinfo{person}{Franco Scarselli}, \bibinfo{person}{Marco
  Gori}, \bibinfo{person}{Ah~Chung Tsoi}, \bibinfo{person}{Markus
  Hagenbuchner}, {and} \bibinfo{person}{Gabriele Monfardini}.}
  \bibinfo{year}{2008}\natexlab{}.
\newblock \showarticletitle{The graph neural network model}.
\newblock \bibinfo{journal}{\emph{IEEE transactions on neural networks}}
  \bibinfo{volume}{20}, \bibinfo{number}{1} (\bibinfo{year}{2008}),
  \bibinfo{pages}{61--80}.
\newblock


\bibitem[Shi et~al\mbox{.}(2020)]%
        {spmm-shi}
\bibfield{author}{\bibinfo{person}{Shaohuai Shi}, \bibinfo{person}{Qiang Wang},
  {and} \bibinfo{person}{Xiaowen Chu}.} \bibinfo{year}{2020}\natexlab{}.
\newblock \showarticletitle{Efficient Sparse-Dense Matrix-Matrix Multiplication
  on {GPUs} Using the Customized Sparse Storage Format}. In
  \bibinfo{booktitle}{\emph{2020 IEEE 26th International Conference on Parallel
  and Distributed Systems (ICPADS)}}. \bibinfo{pages}{19--26}.
\newblock
\urldef\tempurl%
\url{https://doi.org/10.1109/ICPADS51040.2020.00013}
\showDOI{\tempurl}


\bibitem[Shi and Rajkumar(2020)]%
        {Point-GNN}
\bibfield{author}{\bibinfo{person}{Weijing Shi} {and}
  \bibinfo{person}{Ragunathan~(Raj) Rajkumar}.}
  \bibinfo{year}{2020}\natexlab{}.
\newblock \showarticletitle{{Point-GNN}: Graph Neural Network for {3D} Object
  Detection in a Point Cloud}. In \bibinfo{booktitle}{\emph{The IEEE Conference
  on Computer Vision and Pattern Recognition (CVPR)}}.
\newblock


\bibitem[Solomonik et~al\mbox{.}(2017)]%
        {solomonik2017scaling}
\bibfield{author}{\bibinfo{person}{Edgar Solomonik}, \bibinfo{person}{Maciej
  Besta}, \bibinfo{person}{Flavio Vella}, {and} \bibinfo{person}{Torsten
  Hoefler}.} \bibinfo{year}{2017}\natexlab{}.
\newblock \showarticletitle{Scaling betweenness centrality using
  communication-efficient sparse matrix multiplication}. In
  \bibinfo{booktitle}{\emph{ACM/IEEE Supercomputing}}.
\newblock


\bibitem[V{\'a}zquez et~al\mbox{.}(2012)]%
        {spmm-vazquez}
\bibfield{author}{\bibinfo{person}{Francisco V{\'a}zquez},
  \bibinfo{person}{Gloria~Ortega L{\'o}pez},
  \bibinfo{person}{Jos{\'e}-Jes{\'u}s Fern{\'a}ndez},
  \bibinfo{person}{Inmaculada Garc{\'i}a}, {and} \bibinfo{person}{Ester~M.
  Garz{\'o}n}.} \bibinfo{year}{2012}\natexlab{}.
\newblock \showarticletitle{Fast Sparse Matrix Matrix Product Based on {ELLR-T}
  and {GPU} Computing}.
\newblock \bibinfo{journal}{\emph{2012 IEEE 10th International Symposium on
  Parallel and Distributed Processing with Applications}}
  (\bibinfo{year}{2012}), \bibinfo{pages}{669--674}.
\newblock


\bibitem[Veli{\v{c}}kovi{\'c} et~al\mbox{.}(2017)]%
        {gat}
\bibfield{author}{\bibinfo{person}{Petar Veli{\v{c}}kovi{\'c}},
  \bibinfo{person}{Guillem Cucurull}, \bibinfo{person}{Arantxa Casanova},
  \bibinfo{person}{Adriana Romero}, \bibinfo{person}{Pietro Lio}, {and}
  \bibinfo{person}{Yoshua Bengio}.} \bibinfo{year}{2017}\natexlab{}.
\newblock \showarticletitle{Graph attention networks}.
\newblock \bibinfo{journal}{\emph{arXiv preprint arXiv:1710.10903}}
  (\bibinfo{year}{2017}).
\newblock


\bibitem[Wu et~al\mbox{.}(2020)]%
        {wu2020comprehensive}
\bibfield{author}{\bibinfo{person}{Zonghan Wu} {et~al\mbox{.}}}
  \bibinfo{year}{2020}\natexlab{}.
\newblock \showarticletitle{A comprehensive survey on graph neural networks}.
\newblock \bibinfo{journal}{\emph{IEEE Transactions on Neural Networks and
  Learning Systems}} (\bibinfo{year}{2020}).
\newblock


\bibitem[Xie et~al\mbox{.}(2022)]%
        {graphiler}
\bibfield{author}{\bibinfo{person}{Zhiqiang Xie}, \bibinfo{person}{Minjie
  Wang}, \bibinfo{person}{Zihao Ye}, \bibinfo{person}{Zheng Zhang}, {and}
  \bibinfo{person}{Rui Fan}.} \bibinfo{year}{2022}\natexlab{}.
\newblock \showarticletitle{Graphiler: Optimizing Graph Neural Networks with
  Message Passing Data Flow Graph}. In \bibinfo{booktitle}{\emph{Proceedings of
  Machine Learning and Systems}},
  \bibfield{editor}{\bibinfo{person}{D.~Marculescu}, \bibinfo{person}{Y.~Chi},
  {and} \bibinfo{person}{C.~Wu}} (Eds.), Vol.~\bibinfo{volume}{4}.
  \bibinfo{pages}{515--528}.
\newblock
\urldef\tempurl%
\url{https://proceedings.mlsys.org/paper/2022/file/a87ff679a2f3e71d9181a67b7542122c-Paper.pdf}
\showURL{%
\tempurl}


\bibitem[Xu et~al\mbox{.}(2019)]%
        {gin}
\bibfield{author}{\bibinfo{person}{Keyulu Xu}, \bibinfo{person}{Weihua Hu},
  \bibinfo{person}{Jure Leskovec}, {and} \bibinfo{person}{Stefanie Jegelka}.}
  \bibinfo{year}{2019}\natexlab{}.
\newblock \bibinfo{title}{How Powerful are Graph Neural Networks?}
\newblock
\newblock
\showeprint[arxiv]{1810.00826}~[cs.LG]


\bibitem[Yang et~al\mbox{.}(2016)]%
        {dataset-planetoid}
\bibfield{author}{\bibinfo{person}{Zhilin Yang}, \bibinfo{person}{William~W.
  Cohen}, {and} \bibinfo{person}{Ruslan Salakhutdinov}.}
  \bibinfo{year}{2016}\natexlab{}.
\newblock \showarticletitle{Revisiting Semi-Supervised Learning with Graph
  Embeddings}.
\newblock \bibinfo{journal}{\emph{CoRR}}  \bibinfo{volume}{abs/1603.08861}
  (\bibinfo{year}{2016}).
\newblock
\showeprint[arXiv]{1603.08861}
\urldef\tempurl%
\url{http://arxiv.org/abs/1603.08861}
\showURL{%
\tempurl}


\bibitem[Ye et~al\mbox{.}(2023)]%
        {sparsetir}
\bibfield{author}{\bibinfo{person}{Zihao Ye}, \bibinfo{person}{Ruihang Lai},
  \bibinfo{person}{Junru Shao}, \bibinfo{person}{Tianqi Chen}, {and}
  \bibinfo{person}{Luis Ceze}.} \bibinfo{year}{2023}\natexlab{}.
\newblock \showarticletitle{{SparseTIR}: Composable Abstractions for Sparse
  Compilation in Deep Learning}. In \bibinfo{booktitle}{\emph{Proceedings of
  the 28th ACM International Conference on Architectural Support for
  Programming Languages and Operating Systems, Volume 3}} (Vancouver, BC,
  Canada) \emph{(\bibinfo{series}{ASPLOS 2023})}.
  \bibinfo{publisher}{Association for Computing Machinery},
  \bibinfo{address}{New York, NY, USA}, \bibinfo{pages}{660–678}.
\newblock
\showISBNx{9781450399180}
\urldef\tempurl%
\url{https://doi.org/10.1145/3582016.3582047}
\showDOI{\tempurl}


\bibitem[Zeng et~al\mbox{.}(2020)]%
        {dataset-flickr}
\bibfield{author}{\bibinfo{person}{Hanqing Zeng}, \bibinfo{person}{Hongkuan
  Zhou}, \bibinfo{person}{Ajitesh Srivastava}, \bibinfo{person}{Rajgopal
  Kannan}, {and} \bibinfo{person}{Viktor Prasanna}.}
  \bibinfo{year}{2020}\natexlab{}.
\newblock \bibinfo{title}{{GraphSAINT}: Graph Sampling Based Inductive Learning
  Method}.
\newblock
\newblock
\showeprint[arxiv]{1907.04931}~[cs.LG]


\bibitem[Zhang et~al\mbox{.}(2021)]%
        {dgnn}
\bibfield{author}{\bibinfo{person}{Hengrui Zhang}, \bibinfo{person}{Zhongming
  Yu}, \bibinfo{person}{Guohao Dai}, \bibinfo{person}{Guyue Huang},
  \bibinfo{person}{Yufei Ding}, \bibinfo{person}{Yuan Xie}, {and}
  \bibinfo{person}{Yu Wang}.} \bibinfo{year}{2021}\natexlab{}.
\newblock \bibinfo{title}{Understanding {GNN} Computational Graph: A
  Coordinated Computation, {IO}, and Memory Perspective}.
\newblock
\newblock
\showeprint[arxiv]{2110.09524}~[cs.LG]


\bibitem[Zhang et~al\mbox{.}(2020)]%
        {zhang2020deep}
\bibfield{author}{\bibinfo{person}{Ziwei Zhang}, \bibinfo{person}{Peng Cui},
  {and} \bibinfo{person}{Wenwu Zhu}.} \bibinfo{year}{2020}\natexlab{}.
\newblock \showarticletitle{Deep learning on graphs: A survey}.
\newblock \bibinfo{journal}{\emph{IEEE Transactions on Knowledge and Data
  Engineering}} (\bibinfo{year}{2020}).
\newblock


\bibitem[Zhou et~al\mbox{.}(2020)]%
        {zhou2020graph}
\bibfield{author}{\bibinfo{person}{Jie Zhou} {et~al\mbox{.}}}
  \bibinfo{year}{2020}\natexlab{}.
\newblock \showarticletitle{Graph neural networks: A review of methods and
  applications}.
\newblock \bibinfo{journal}{\emph{AI Open}}  \bibinfo{volume}{1}
  (\bibinfo{year}{2020}), \bibinfo{pages}{57--81}.
\newblock


\end{thebibliography}
}

\appendix

\section*{Appendix}

\section{No memory overhead with GCN caching}
\label{sec:no-mem-overhead}

\begin{figure}
    \begin{subfigure}{0.89\textwidth}
        \centering
        \includegraphics[clip, trim=0cm 0cm 34cm 11cm, width=1.00\textwidth]{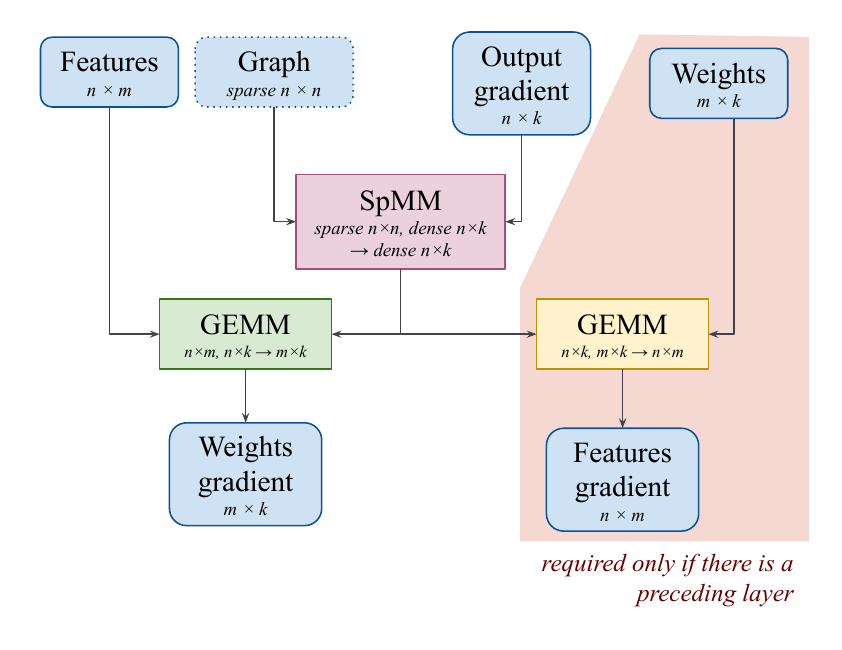}
        \caption{Fused-propagate GCN backward pass, one $n \times k$ SpMM.}
        \label{subfig:gcn-bwd-reuse}
    \end{subfigure}
    \begin{subfigure}{0.92\textwidth}
        \centering
        \includegraphics[clip, trim=15cm 0cm 16.5cm 11cm, width=1.00\textwidth]{figures/GNN_thesis_GCN_order_full.pdf}
        \caption{Split-propagate GCN backward pass, two $n \times m$ SpMMs.}
        \label{subfig:gcn-bwd-2spmm}
    \end{subfigure}
    \caption{Alternative schemes of computing the backward pass for GCN. Red area on the right indicates part of computation that does not need to be executed if the feature gradients are not needed. Compute nodes of the same color operate on the same shapes, which are indicated on the node in Einstein summation notation. }
    \label{fig:gcn-bwd-order}
\end{figure}

\begin{figure}
    \begin{subfigure}{\textwidth}
        \centering
        \includegraphics[clip, trim=32cm 12cm 5cm 0cm, width=0.6\textwidth]{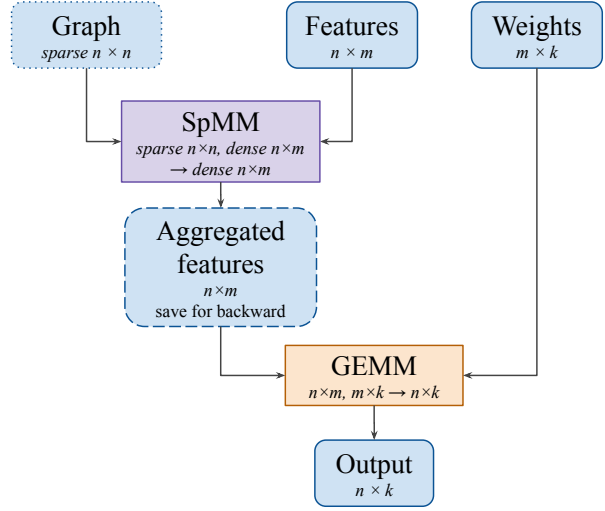}
        \caption{GCN forward pass with caching. One $n \times k$ SpMM.}
        \label{subfig:gcn-fwd-cache}
    \end{subfigure}
    \begin{subfigure}{\textwidth}
        \centering
        \includegraphics[clip, trim=32cm 0cm 0cm 11cm, width=1.00\textwidth]{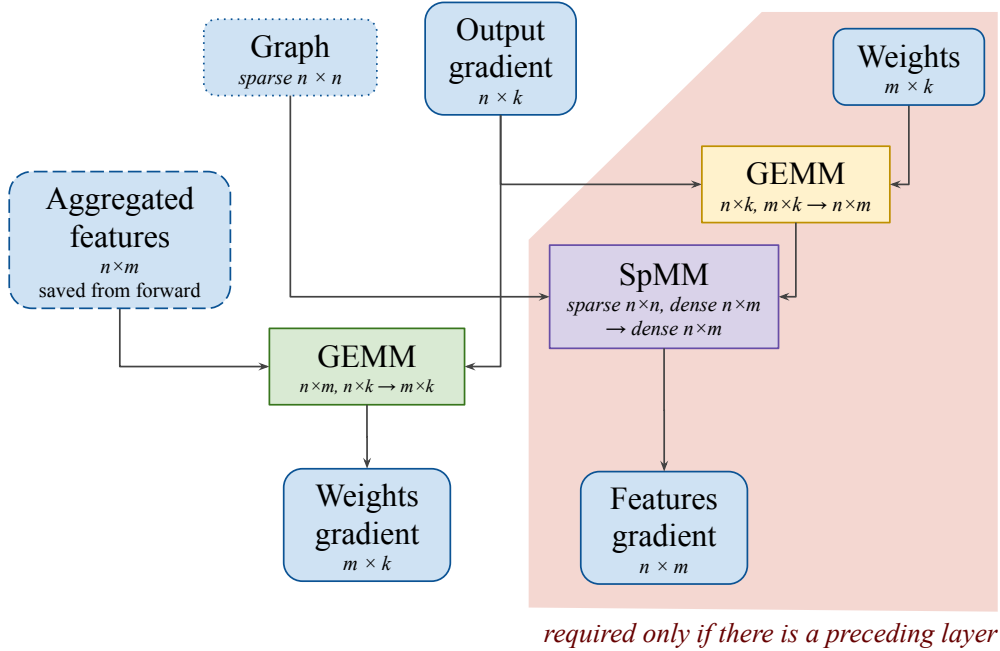}
        \caption{GCN backward pass using cached values. One $n \times m$ SpMM.}
        \label{subfig:gcn-bwd-cache}
    \end{subfigure}
    \caption{GCN computation scheme using caching to avoid recalculation. Red area on the right indicates part of computation that does not need to be executed if the feature gradients are not needed. Compute nodes of the same color operate on the same shapes, which are indicated on the node in Einstein summation notation. }
    \label{fig:gcn-scheme-cache}
\end{figure}

Usually, caching additional values for the backward pass means using additional memory. However, that is not the case when employing caching in GCN. Looking at Figure \ref{fig:gcn-bwd-order}, it can be seen that in order to compute the gradients without caching, in both schemes we need to store the layer input features of size $n \times m$ for the backward pass. Nevertheless, as can be seen from Figure \ref{subfig:gcn-bwd-cache}, once the cached aggregated features are used, the original layer input features are not needed. Thus, the amount of memory used to store intermediates for the backward pass is exactly the same as in the scheme without caching.

\section{GAT Operator Reordering}
\label{sec:gat-op-reord}

For each edge, we need to compute the raw attention weight 
$\mathbf{a}^T[\mathbf{x}_i\mathbf{\Theta} \, \Vert \, \mathbf{x}_j\mathbf{\Theta}]$ 
where $\Vector{a} \in \R^{2k}$, $x_i\Mat{\Theta} \in \R^k$.
If we were to compute that expression directly for each edge, the computational complexity would be $\bigO(qk)$ where $q$ is the number of edges.

However, there exists a much more efficient scheme that is widely employed by other works \cite{pyg, dgnn} and mainstream GNN frameworks.
Instead of computing the whole expression for each edge, we precompute values for each node and only sum them up for each edge.

Firstly, we split the attention weights into two separate vectors: 
$$\Vector{a}~=~\begin{bmatrix}\Vector{a}_\src \\ \Vector{a}_\dst\end{bmatrix}$$

where $\Vector{a}_{\text{src}}, \Vector{a}_{\text{dst}} \in \R^k$. 
This allows us to first compute 
$\Matrix{Y} = \Matrix{X} \Matrix{\Theta}$ 
and then for each node $i$ calculate 
$\alpha^\src = \Vector{a}_\src^T \Matrix{Y}_i$ 
and 
$\alpha^\dst=\Vector{a}_\dst^T\Matrix{Y}_i$. 
Then for each edge between nodes $i$ and $j$ we only need to sum $\alpha^\src_i + \alpha^\dst_j$. 
This way of computing the weights results in computational complexity of $\bigO(q + nk)$ which is much more beneficial, given that $n \ll q$ in graph data.

\section{GAT Caching}
\label{sec:gat-caching}

\begin{figure}[p]
    \centering
    \includegraphics[clip, trim=0cm 9cm 27.5cm 0cm, width=\textwidth]{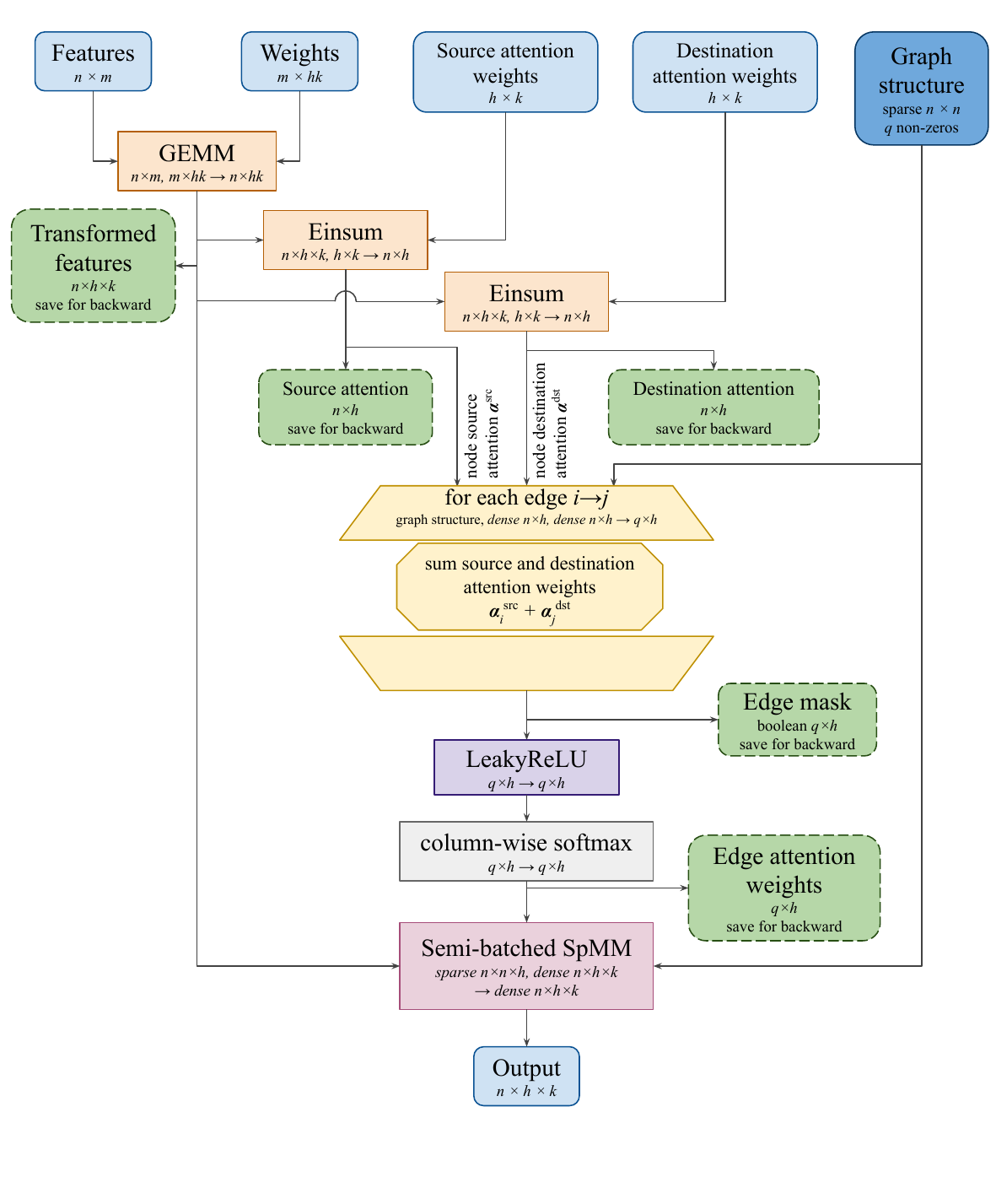}
    \caption{GAT forward scheme. Green boxes with dashed frames indicate values that can be cached for the backward pass. }
    \label{fig:gat-scheme-cache}
\end{figure}

\begin{figure}[p]
    \centering
    \includegraphics[clip, trim=26.5cm 0cm 0cm 0cm, width=0.98\textwidth]{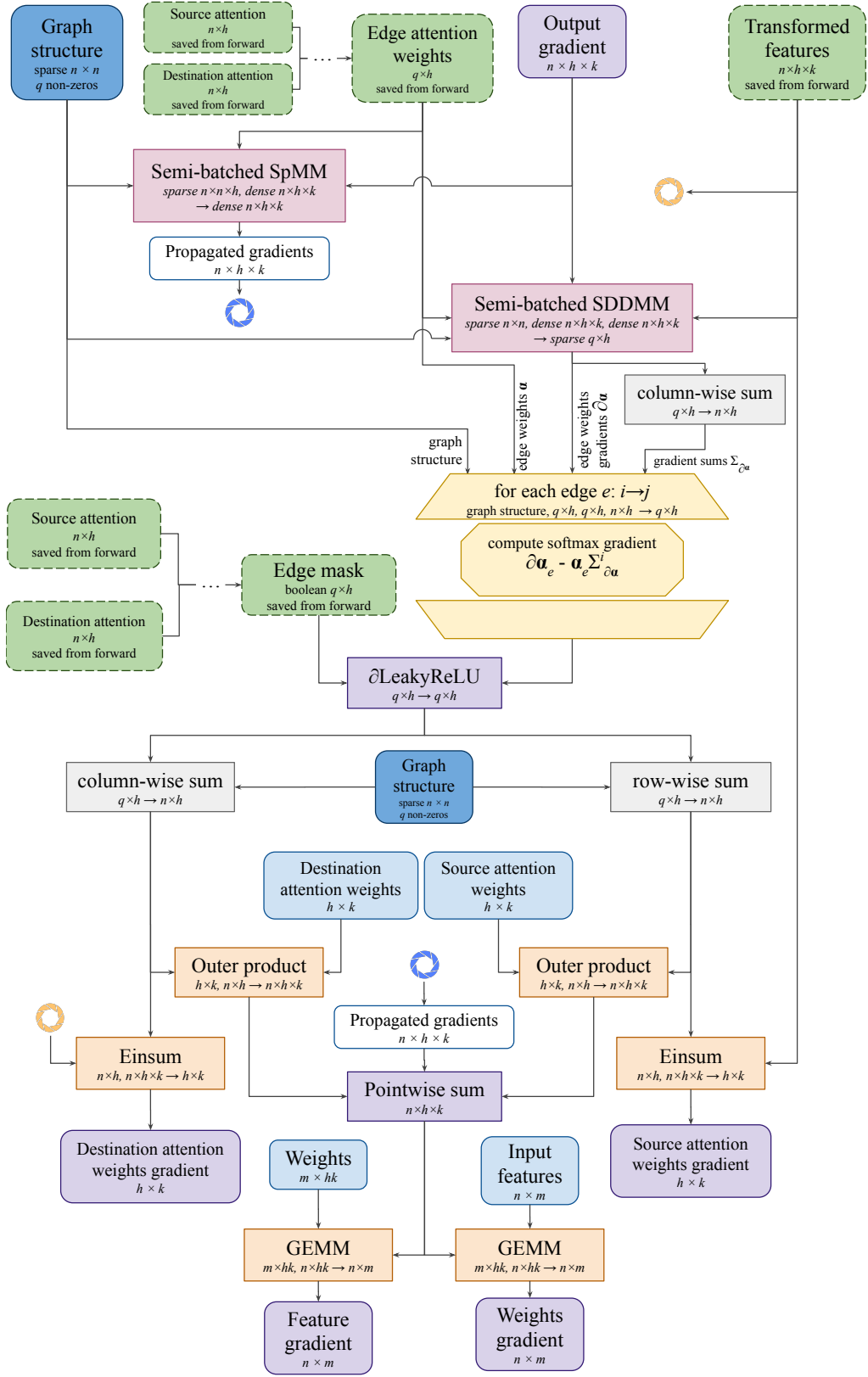}
    \caption{GAT backward pass using cached values. Source and destination attention can be saved instead of edge mask and edge attention weights to use less memory. The scheme for recomputation of values from the forward pass is omitted for clarity.}
    \label{fig:gat-bwd-cache}
\end{figure}

Similar to GCN, in GAT we can cache intermediate values to avoid recomputing them in the backward pass. As can be seen in Figure \ref{fig:gat-scheme-cache}, there are multiple stages at which we can cache intermediate values. In Figure \ref{fig:gat-bwd-cache} the scheme for computing the backward pass can be seen.

\section{Sparse Formats}
\label{sec:sparse-formats}

\begin{figure}
    \centering
    \includegraphics[width=0.95\textwidth]{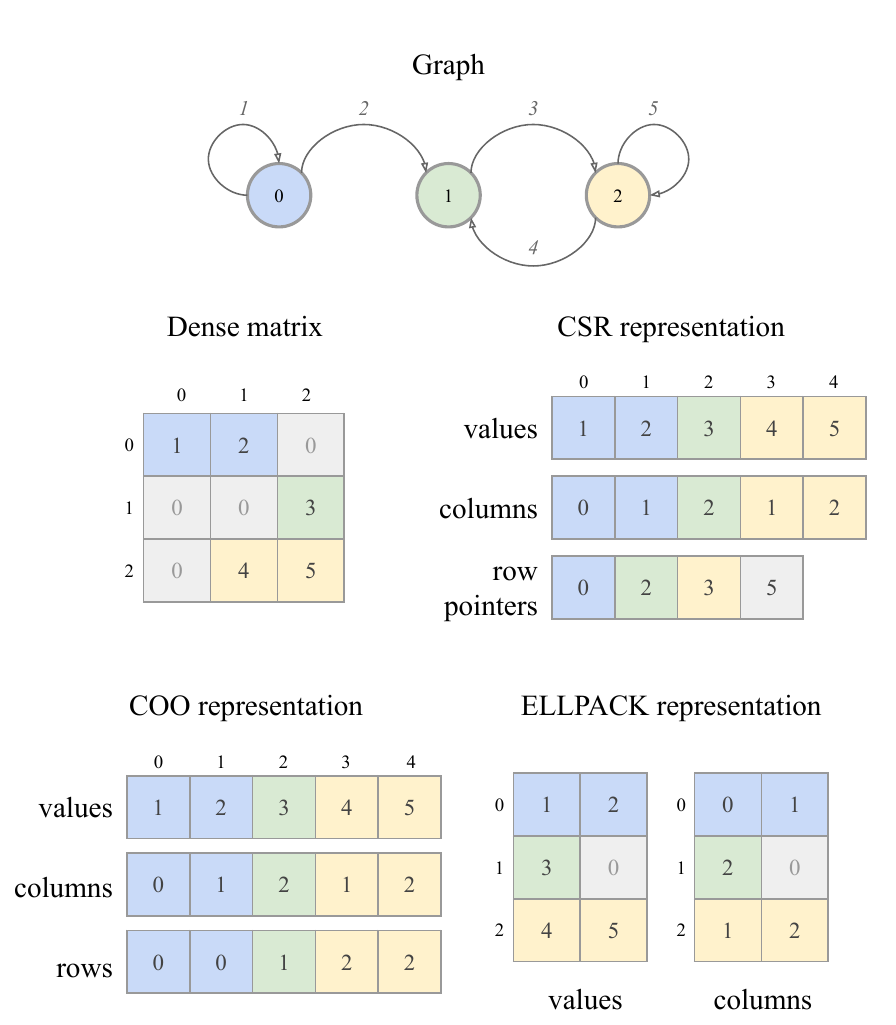}
    \caption{Example graph and its representations in selected formats.}
    \label{fig:sparse-formats}
\end{figure}

Let us assume a matrix $\Matrix{A} \in \R^{n \times m}$. All of the formats mentioned below are illustrated in Figure \ref{fig:sparse-formats}. 

\subsection{Compressed Sparse Row and Compressed Sparse Column}
The Compressed Sparse Row format (CSR) is one of the most commonly used sparse matrix formats. 

CSR requires storing two vectors describing the sparse structure: the columns array $\Vector{c} \in \Z^q$ and the row pointer array $\Vector{r} \in \Z^{n+1}$. The $i$-th position in $\Vector{r}$ indicates at which index of $\Vector{c}$ the $i$-th row start. The $(n+1)$-th position pf $\Vector{r}$ contains the value $q+1$ to simplify array processing. The $j$-th position in $\Vector{c}$ indicates the column of the $j$-th value in the dense representation of the matrix. Optionally, a third array can be used to store edge weights: the values array $\Vector{v} \in \R^q$, which is indexed the same way as $\Vector{c}$. 
In total, this format requires the storage of $q + n + 1$ integer values for indexing and $q$ floating point values to represent the non-zero matrix entries. Thus, among the formats described here, the CSR format requires the least memory to store a given sparse matrix.

The CSR format is not particularly well suited to processing on GPUs because it enforces row-wise processing of the matrix but the rows are not even. Thus, the work division between CUDA threads is not straightforward and requires dynamic load balancing.

The Compressed Sparse Column format (CSC) is analogous to the CSR format. Instead of the arrays representing columns and row pointers, it uses one array for rows and one for column pointers.

\subsection{Coordinate Format}

The coordinate format (COO) is the simplest sparse matrix format. To represent a matrix in this format, two arrays are needed: the columns array $\Vector{c} \in \Z^q$ and the rows array $\Vector{r} \in \Z^{q}$, that together indicate the coordinates of a given entry in the original dense matrix. Similarly to CSR, a third array storing values can be also used: the values array $\Vector{v} \in \R^q$. 

This format has a higher memory requirement than CSR and CSC. It requires storing three matrices of length $q$ each (two holding integer values and one holding floating point values). However, it allows for a simpler parallelization. The matrix can be processed entry-by-entry, so no tailored scheme is needed to ensure even work division between threads.

\subsection{ELLPACK Format}

The ELLPACK format %
is better suited for processing on GPUs than CSR and COO. Instead of one-dimensional vectors, it uses a two-dimensional matrix to store the sparsity structure: $\Matrix{Col} = (c_{ij}) \in \Z^{n \times p}$, where $c_{ij}$ holds the column index of the $j$-th entry in the $i$-th row. The number of columns in $\Matrix{Col}$, which we denote by $p$, is equal to the highest number of non-zero entries in a single row of the matrix.
Similarly, to store the non-zero matrix values, a two-dimensional matrix is used: the values matrix $\Matrix{V} = (v_{ij}) \in \R^{n \times p}$, where $v_{ij}$ stores the value of the~$j$-th entry in the $i$-th row.

In ELLPACK, a more regular representation is obtained at the cost of storing unnecessary data. This format requires storing $np$ integer values for indexing and another $np$ floating point values. Thus, the ELLPACK format is well-suited to processing of matrices that have non-zero values evenly distributed across the rows. Otherwise, a single row that is much longer than the others leads to unnecessarily high storage needs. Due to this, ELLPACK is not well suited for certain graphs, such as graphs with hubs or other subgraphs that imply high variance in node degree. 
However, the regular structure of the ELLPACK representation makes it easy to process in parallel on GPUs.

\subsection{Hybrid Formats} \label{sec:hybrid-formats}

As an attempt to mitigate the disadvantages stemming from the use of some of the formats, a hybrid format can be used. There are multiple variants, such as hybrid formats used by Ye et al. \cite{sparsetir}. As an example a CSR-COO data format could be constructed to mitigate the issues from irregular row lengths in CSR. In this format, only the first $t$ elements in a given row are stored in the CSR format, while the remaining entries are stored in the COO format. 

\section{Common Sparse Operators}

GNNs are characterized by a substantial amount of sparse computation. There are two sparse operators that are particularly common in GNNs: sparse matrix-matrix multiplication and sampled dense-dense matrix multiplication.
  
\subsection{Sparse Matrix-Matrix Multiplication}
The \emph{sparse matrix-matrix multiplication} operator (SpMM) is a multiplication of a sparse matrix $\Matrix{A} \in \R^{n \times m} $ and a dense matrix $\Matrix{B} \in \R^{m \times k}$, resulting in a dense matrix $\Matrix{C} \in \R^{n \times k}$, $\Matrix{C} = \Matrix{A} \Matrix{B}$. 
There exist highly optimized implementations of this operator \cite{cusparse} and it is widely explored in literature \cite{spmm-shi, spmm-vazquez} (see more in \Cref{sec:related-work}). An example implementation using the CSR format can be seen in Listing \ref{lst:spmm-csr}.

\begin{minipage}{\linewidth}
\begin{lstlisting}[language=Python, caption=Example SpMM implementation in NumPy using the CSR format., label={lst:spmm-csr}]
def spmm(A_rowptrs, A_columns, A_values, B):
    M, K = B.shape
    N = A_rowptrs.shape[0] - 1
    C = np.empty((N, K))
    for i in range(N):
        for j in range(A_rowptrs[i], A_rowptrs[i + 1]):
            column = A_columns[j]
            for k in range(K):
                C[i, k] += B[column, k] * A_values[j]
    return C
\end{lstlisting}
\end{minipage}

The SpMM operator is a very common operator in GNNs. It represents the propagation of information between nodes. %
To understand the computational characteristics of SpMM, we look at \emph{operational intensity}, defined as the number of floating point operations executed by the program per byte of I/O. 

Operational intensity for SpMMs executed in GNNs can be found in Table~\ref{tab:spmm-op-intensity}. It can be seen that the operational intensity is no higher than $1.066 \fpb$ which indicates that the computation is memory-bound. Operational intensities of CSC-SpMM and CSR-SpMM are the same because the sparse matrix is square.
Moreover, it is worth noting that ELLPACK is well-suited to represent typical GNN dataset such as Cora and Arxiv. They both have nodes with much higher degrees than average: for Cora, the maximum node degree is 168 and the average degree is 7.8, while for Arxiv the maximum degree is 436 and the average degree is 13.77. Therefore, their ELLPACK value matrices consist in $>96\%$ of zeros, resulting in significantly lower operational intensity when executing SpMM in comparison to other mentioned formats.

\begin{table}[]
\footnotesize
\centering
\begin{tabular}{@{}lrrrrr@{}}
\toprule
\multirow{2}{*}{\textbf{\begin{tabular}[c]{@{}l@{}}Data \\ format\end{tabular}}} &
  \multirow{2}{*}{\textbf{Variables}} &
  \multirow{2}{*}{\textbf{FLOPs}} &
  \multirow{2}{*}{\textbf{I/O $[\text{bytes}]$}} &
  \multicolumn{2}{r}{\textbf{Operational intensity $[\fpb]$}} \\ \cmidrule(l){5-6} 
 &
   &
   &
   &
  \multicolumn{1}{c}{\textbf{\hspace{0.8cm}Cora}} &
  \multicolumn{1}{c}{\textbf{OGB Arxiv}} \\ \midrule
CSR &
  \begin{tabular}[c]{@{}r@{}}$\Vector{v} \in \mathbb{R}^{q}$\\ $\Vector{r}_{ptr} \in \mathbb{R}^{n + 1}$\\ $\Vector{c} \in \mathbb{R}^{q}$\\ $\Matrix{B} \in \mathbb{R}^{m \times f}$\end{tabular} &
  $\bigO(qf)$ &
  $\bigO(q + mf + nf)$ &
  \hspace{0.8cm}0.621 &
  1.066 \\ \midrule
CSC &
  \begin{tabular}[c]{@{}r@{}}$\Vector{v} \in \mathbb{R}^{q}$\\ $\Vector{c}_{ptr} \in \mathbb{R}^{m + 1}$\\ $\Vector{r} \in \mathbb{R}^{q}$\\ $\Matrix{B} \in \mathbb{R}^{m \times f}$\end{tabular} &
  $\bigO(qf)$ &
  $\bigO(q + mf + nf)$ &
  \hspace{0.8cm}0.621 &
  1.066 \\ \midrule
COO &
  \begin{tabular}[c]{@{}r@{}}$\Vector{v} \in \mathbb{R}^{q}$\\  $\Vector{r} \in \mathbb{R}^{q}$\\ $\Vector{c} \in \mathbb{R}^{q}$\\ $\Matrix{B} \in \mathbb{R}^{m \times f}$\end{tabular} &
  $\bigO(qf)$ &
  $\bigO(q + mf + nf)$ &
  \hspace{0.8cm}0.612 &
  1.036 \\ \midrule
ELLPACK &
  \begin{tabular}[c]{@{}r@{}}$\Mat{V} \in \mathbb{R}^{n \times p}$\\  $\Mat{Col} \in \mathbb{R}^{n \times p}$\\ $\Matrix{B} \in \mathbb{R}^{m \times f}$\end{tabular} &
  \multicolumn{1}{l}{$\bigO(qf)$} &
  $\bigO(np + mf + nf)$ &
  \hspace{0.8cm}0.236 &
  0.207 \\ \bottomrule
\end{tabular}%
\caption{Operational intensity of SpMM $\Mat{C} = \Mat{A} \cdot \Mat{B}$. Example operational intensity values were computed for scenarios realistic in GNNs, i.e., where $\Mat{A}$ is the adjacency matrix of a given graph and $\Mat{B}$ has $n$ rows and 64 columns. Detailed information on the graphs can be found in Table \ref{tab:graph_datasets}. To count FLOPs for ELLPACK, we considered only operations on non-zero values.}
\label{tab:spmm-op-intensity}
\end{table}

\subsection{Sampled Dense-Dense Matrix Multiplication}

Another operator that can be encountered in GNN computation is the \emph{sampled dense-dense matrix multiplication} (SDDMM). Given a sparse matrix $\Matrix{A} \in \R^{n \times k} $ and two dense matrices $\Matrix{B} \in \R^{n \times m}$, $\Matrix{C} \in \R^{m \times k}$, we can compute $\Matrix{D} = \Matrix{A} \odot (\Matrix{B} \cdot \Matrix{C})$, where $\odot$ represents the Hadamard product and $\Matrix{D} \in \R^{n \times k}$ is a sparse matrix. 
Similarly to SpMM, this subroutine has existing highly optimized implementations \cite{cusparse}. An example implementation of SDDMM can be found in Listing~\ref{lst:sddmm-csr}.

The SDDMM operator is used in the backward pass of GAT. There, we always need to compute it on matrices which have shapes $\Mat{A} \in \R^{n \times n}, \Mat{B}\in\R^{n \times f}$ and $\Mat{C} \in \R^{f \times n}$. In~Table~\ref{tab:sddmm-op-intensity} we present estimated operational intensity of SDDMM when used in GAT backward pass computation.

\begin{minipage}{\linewidth}
\begin{lstlisting}[language=Python, caption=Example SDDMM implementation in NumPy using the CSR format., label={lst:sddmm-csr}]
def sddmm(A_rowptrs, A_columns, A_values, B, C):
    M, K = B.shape
    N = A_rowptrs.shape[0] - 1
    nnz = A_values.shape[0]
    
    D_values = np.zeros((nnz,))
    for i in range(N):
        for j in range(A_rowptrs[i], A_rowptrs[i + 1]):
            column = A_columns[j]
            D_values[j] = A_values[j] * np.dot(B[i], C[column])
    return D_values
\end{lstlisting}
\end{minipage}

Similarly to SpMM, SDDMM is an operator with low operational intensity. For both Cora and Arxiv, the operational intensity is below $0.7$, independent of the data format. Therefore, SDDMM is also memory-bound. Performing the computation using the ELLPACK format leads to even lower operational intensity due to irregular row sizes in the sparse matrices, same as in the case of SpMM.

\begin{table}[]
\centering
\begin{tabular}{@{}lrrrrr@{}}
\toprule
\multirow{2}{*}{\textbf{\begin{tabular}[c]{@{}l@{}}Data \\ format\end{tabular}}} &
  \multirow{2}{*}{\textbf{Variables}} &
  \multirow{2}{*}{\textbf{FLOPs}} &
  \multirow{2}{*}{\textbf{I/O $[\text{bytes}]$}} &
  \multicolumn{2}{r}{\textbf{Operational intensity $[\fpb]$}} \\ \cmidrule(l){5-6} 
 &
   &
   &
   &
  \multicolumn{1}{c}{\textbf{\hspace{0.8cm}Cora}} &
  \multicolumn{1}{c}{\textbf{OGB Arxiv}} \\ \midrule
CSR &
  \begin{tabular}[c]{@{}r@{}}$\Vector{v} \in \mathbb{R}^{q}$\\ $\Vector{r}_{ptr} \in \mathbb{R}^{n + 1}$\\ $\Vector{c} \in \mathbb{R}^{q}$\\ $\Matrix{B} \in \mathbb{R}^{n \times f}$\\ $\Matrix{C} \in \mathbb{R}^{f \times n}$\end{tabular} &
  $\bigO(qf)$ &
  $\bigO(qf+n)$ &
  \hspace{0.8cm}0.567 &
  0.620 \\ \midrule
CSC &
  \begin{tabular}[c]{@{}r@{}}$\Vector{v} \in \mathbb{R}^{q}$\\ $\Vector{c}_{ptr} \in \mathbb{R}^{m + 1}$\\ $\Vector{r} \in \mathbb{R}^{q}$\\ $\Matrix{B} \in \mathbb{R}^{n \times f}$\\ $\Matrix{C} \in \mathbb{R}^{f \times n}$\end{tabular} &
  $\bigO(qf)$ &
  $\bigO(qf+n)$ &
  \hspace{0.8cm}0.567 &
  0.613 \\ \midrule
COO &
  \begin{tabular}[c]{@{}r@{}}$\Vector{v} \in \mathbb{R}^{q}$\\  $\Vector{r} \in \mathbb{R}^{q}$\\ $\Vector{c} \in \mathbb{R}^{q}$\\ $\Matrix{B} \in \mathbb{R}^{n \times f}$\\ $\Matrix{C} \in \mathbb{R}^{f \times n}$\end{tabular} &
  $\bigO(qf)$ &
  $\bigO(qf)$ &
  \hspace{0.8cm}0.562 &
  0.613 \\ \midrule
ELLPACK &
  \begin{tabular}[c]{@{}r@{}}$\Mat{V} \in \mathbb{R}^{n \times p}$\\  $\Mat{Col} \in \mathbb{R}^{n \times p}$\\ $\Matrix{B} \in \mathbb{R}^{n \times f}$\\ $\Matrix{C} \in \mathbb{R}^{f \times n}$\end{tabular} &
  $\bigO(qf)$ &
  $\bigO(npf)$ &
  \multicolumn{1}{c}{\hspace{0.8cm}0.308} &
  0.325 \\ \bottomrule
\end{tabular}%
\caption{Operational intensity of SDDMM $\Matrix{D} = \Matrix{A} \odot (\Matrix{B} \Matrix{C})$. Example operational intensity values were computed for scenarios realistic in GNNs, i.e., where $\Mat{A}$ is the adjacency matrix of a given graph and $\Mat{B}, \Mat{C}$ have $n$ rows and 64 columns. Detailed information on the graphs can be found in Table \ref{tab:graph_datasets}. Counting the FLOPs for ELLPACK, we consider only operations on non-zero values.}
\label{tab:sddmm-op-intensity}
\end{table}

\section{GCN Evaluation}
\label{sec:gcn-eval-details}

\begin{figure} 
    \centering
    \begin{subfigure}{\textwidth}
        \centering
        \includegraphics[width=\textwidth]{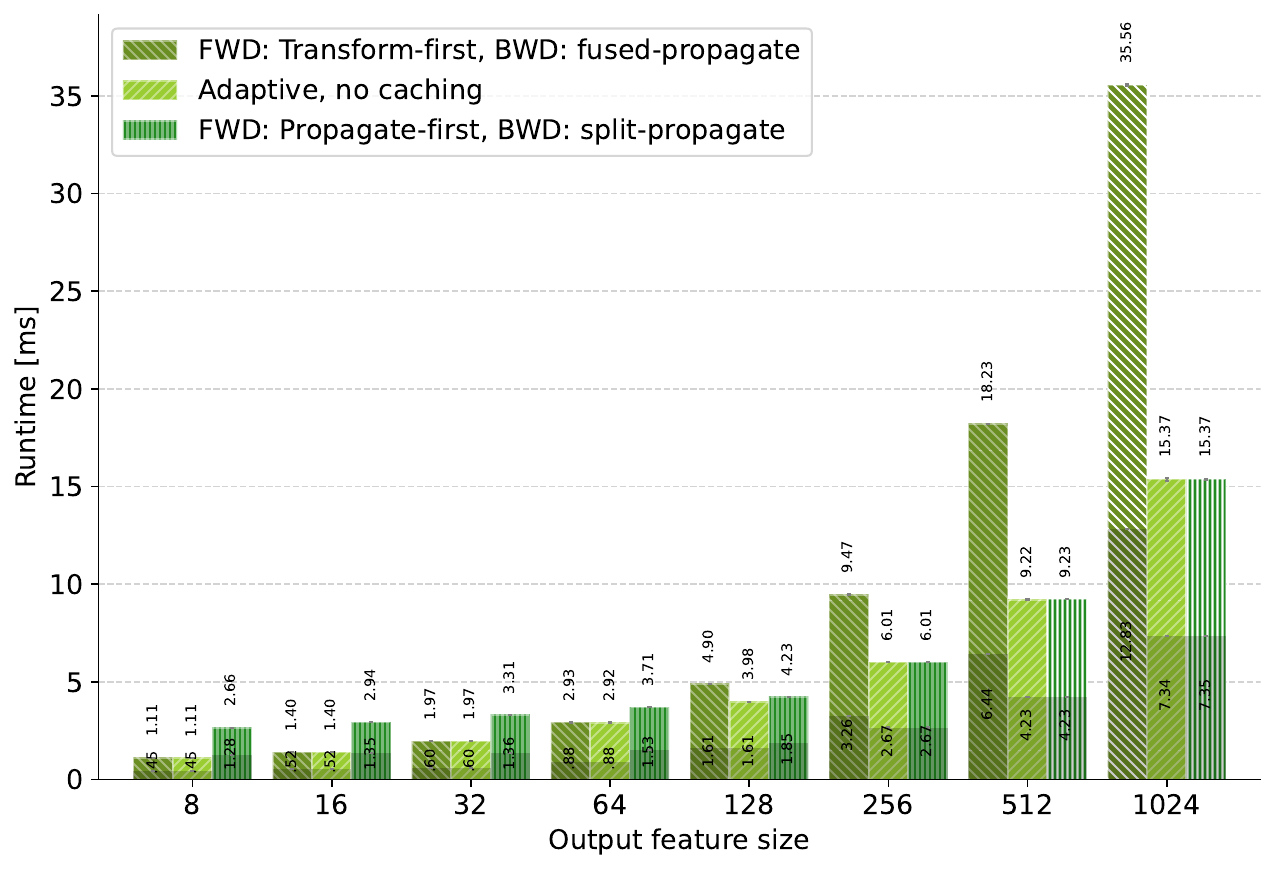}
        \caption{Runtime without calculating input gradients.}
    \end{subfigure}
    \begin{subfigure}{\textwidth}
        \centering
        \includegraphics[width=\textwidth]{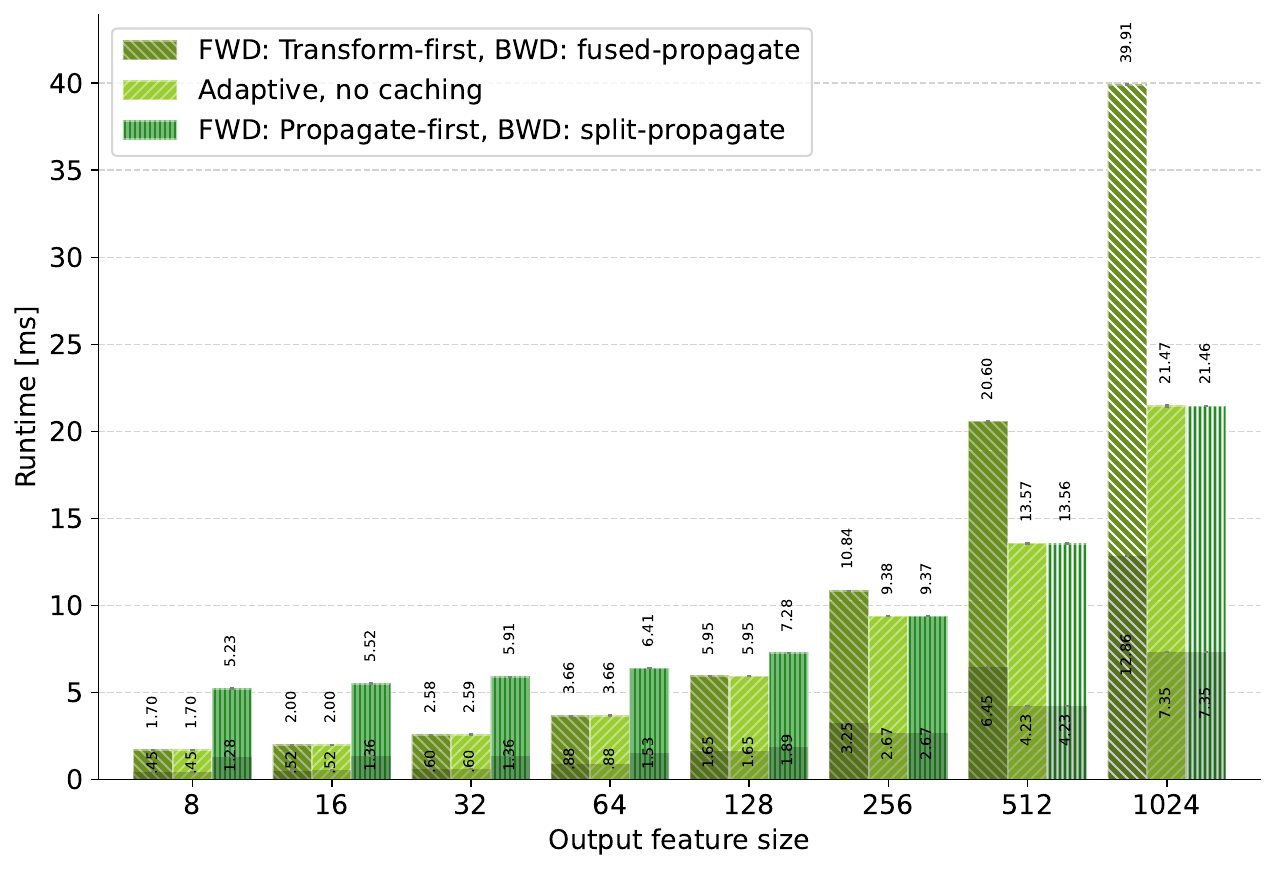}
        \caption{Runtime including calculating input gradients.}
    \end{subfigure}
    \caption{Single GCN layer runtime on the OGB Arxiv dataset. Darker areas represent the time spent in the forward pass, total bar height is the total time of forward and backward passes.}
    \label{fig:exp-gcn-evaluate-schemes}
\end{figure}

The results for schemes without caching are presented in Figure \ref{fig:exp-gcn-evaluate-schemes}. It can be seen that the runtime patterns follow our expectations from Table \ref{tab:gcn-schemes}.
In the case of not computing input feature gradients, as long as the number of input features (128) is bigger than that of the output features, the transform-first forward and fused-propagate backward pass scheme is faster. However, once the number of features surpasses 128, the alternative scheme is faster. 
The same applies to the case with computing input gradients, but the threshold is shifted to 256. Our adaptive implementation is always as fast as the faster scheme. 

\begin{figure}
    \centering

    \begin{subfigure}{\textwidth}
        \centering
        \includegraphics[width=\textwidth]{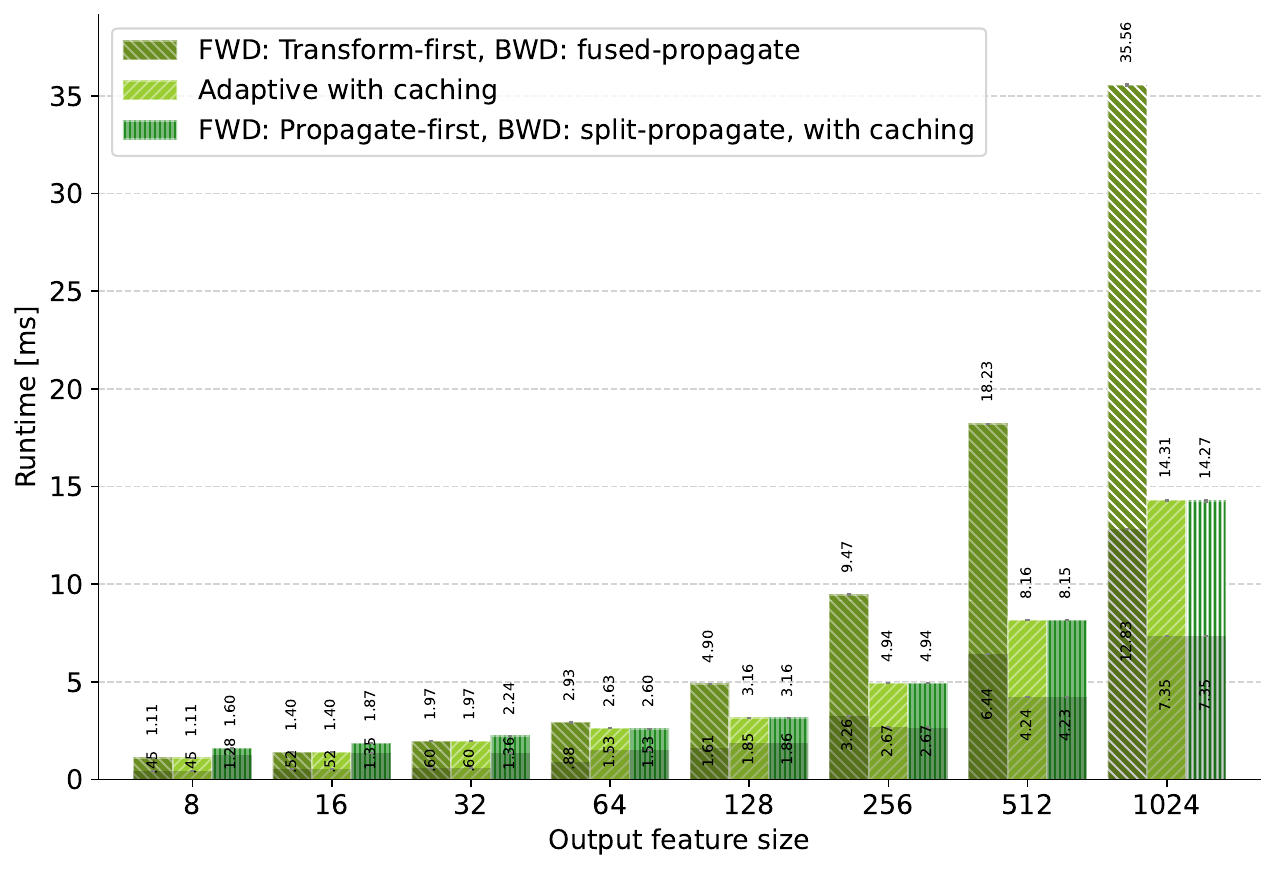}
        \caption{Runtime without calculating input gradients.}
    \end{subfigure}
    \begin{subfigure}{\textwidth}
        \centering
        \includegraphics[width=\textwidth]{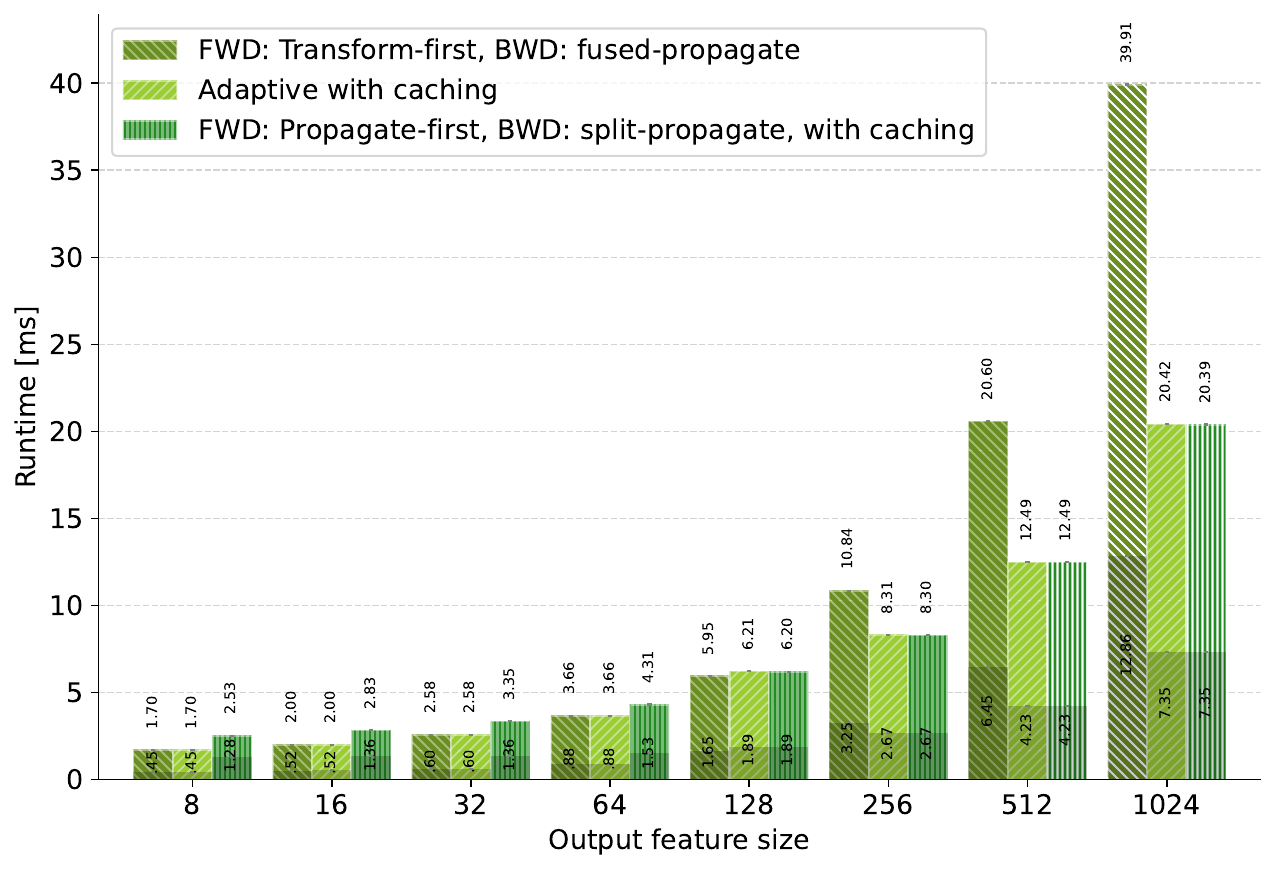}
        \caption{Runtime including calculating input gradients.}
        \label{subfig:gcn-inner-inp-grad}
    \end{subfigure}
    
    \caption{Single GCN layer runtime on the OGB Arxiv dataset \emph{using caching}. Darker areas represent the time spent in the forward pass, total bar height is the total time of forward and backward passes.}
    \label{fig:exp-gcn-caching-schemes}
\end{figure}

The results for schemes with caching are presented in Figure \ref{fig:exp-gcn-caching-schemes}. It can be seen that the runtime distribution follows what we expected in Table \ref{tab:gcn-cached-schemes}. When not computing input gradients, the scheme using caching is faster once the number of output features exceeds half of the input features.
When calculating input gradients, the threshold for adapting the scheme lies at equal input and output feature sizes, which is 128. Again, our adaptive computing scheme proves to match the fastest scheme in every case except the threshold point. 

The single suboptimal choice of the scheme at the hidden size of 128 in Figure \ref{subfig:gcn-inner-inp-grad} is due to the fact that we assumed that running two SpMMs matrices matrices of sizes $n \times n$ and $n \times t$, where $t \in \Z^+$ would take the same time as running one SpMM multiplying matrices of sizes $n \times n$ and $n \times 2t$ (limitations of our approach are described in \Cref{sec:gcn-analysis}).

\section{Implementations}
\label{sec:gcn-gat-implementaitons}

\subsection{GCN Operator Implementation}
\begin{lstlisting}[language=Python, caption=Example GCN operator implementation using NumPy., label={lst:gcn-fwd}]
def gcn(node_features, rowptrs, columns, edge_vals, weights, output):
    # GEMM
    new_features = node_features @ weights

    # SpMM
    output[:] = 0
    for i, k in dace.map[0:N, 0:num_out_features]:
        for j in dace.map[rowptrs[i]:rowptrs[i + 1]]:
            column = columns[j]
            mult = new_features[i, k] * vals[j]
            output[column, k] += mult
\end{lstlisting}

\subsection{GCN Forward Pass}
\label{sec:gcn-fwd}

More specifically, the GCN operator takes as input an \textit{adjacency matrix} $\mathbf{A} \in \mathbb{R}^{n \times n}$ and \textit{node features} $\mathbf{X} \in \mathbb{R}^{n \times m}$. 

The resulting node features $\mathbf{X'} \in \mathbb{R}^{n \times k}$ are computed as follows:

\begin{equation}
    \mathbf{X'} = \mathbf{D}^{-1/2} \mathbf{A} \mathbf{D}^{-1/2} \mathbf{X} \mathbf{\Theta} + \mathbb{1}_{n}\mathbf{b}^T
\end{equation}
where $\mathbf{D} \in \mathbb{R}^{n \times n}$ is the diagonal degree matrix, $\mathbf{\Theta} \in \R^{m \times k}$ is the learned parameter matrix, $\mathbf{b} \in \mathbb{R}^k$ is the learned bias vector, and $\mathbb{1}_{n}$ is an $n$-element vector of ones.

Optionally, edge weights can be included as entries other than $1$ in the adjacency matrix.
Often, self-loops are also inserted, so a modified adjacency matrix is used: $\mathbf{\hat{A}} = \mathbf{A} + \mathbf{I}$. 
As the graphs are constant during training, the data can be preprocessed as following:

\begin{equation}
    \mathbf{A'} = \hat{\mathbf{D}}^{-1/2}\mathbf{\hat{A}}\hat{\mathbf{D}}^{-1/2}
\end{equation}

where $\hat{\mathbf{D}}$ is the diagonal degree matrix for a graph with added self-loops. Then, the GCN formulation becomes:

\begin{equation} \label{eq:gcn-fwd}
    \mathbf{X'} = \mathbf{A'} \mathbf{X} \mathbf{\Theta} + \mathbb{1}_{n}\mathbf{b}^T
\end{equation}

where $\Matrix{A'}$ is a sparse matrix, while all others are dense.

Above formulation assumes the use of \texttt{sum} as the message aggregation function, so the message passing and aggregation are represented by the matrix multiplication $\Mat{A} \cdot \Mat{X} \Mat{\Theta}$. Some models alternatively use \texttt{mean} as the aggregation function, which can be represented by $\Mat{D} \cdot \Mat{A} \cdot \Mat{X} \Mat{\Theta}$. Another alternative aggregation function is \texttt{max}, which can be represented in the node-wise formulation as following:

$$
x_{ik} = \max_{j \in \mathcal{N}(i)} \left( \Vector{x_j}\Matrix{\Theta}_k^T\right)
$$

Throughout this work we focus on networks with the aggregation function \texttt{sum}.  

\subsection{GAT Forward Pass}
\label{sec:gat-fwd}

The GAT operator can be computed as follows:

\begin{equation}\label{eq:gat-fwd}
    \mathbf{X'} = \mathcal{A} \Matrix{X} \Matrix{\Theta} + \mathbb{1}_{n}\mathbf{b}^T,
\end{equation}
where $\mathcal{A} = (\alpha_{ij})$ is the sparse attention weight matrix, $\mathbf{\Theta} \in \mathbb{R}^{m \times k}$ is the learned parameter matrix, $\mathbf{b} \in \mathbb{R}^k$ is the learned bias vector.

The attention weight of the edge from node $i$ to node $j$ is defined as:

\begin{equation}\label{eq:gat-fwd-att}
\alpha_{ij} =
        \frac{
        \exp\left(f\left(\mathbf{a}^T
        [\mathbf{\Theta}^T\mathbf{x}_i \, \Vert \, \mathbf{\Theta}^T \mathbf{x}_j]
        \right)\right)}
        {\sum_{k \in \mathcal{N}(i) \cup \{ i \}}
        \exp\left(f\left(\mathbf{a}^T
        [\mathbf{\Theta}^T\mathbf{x}_i \, \Vert \, \mathbf{\Theta}^T \mathbf{x}_k]
        \right)\right)},
\end{equation}

where $\Matrix{X} \in \R^{n \times m}$ is the input node features matrix, $||$ is the concatenation operator, $\mathbf{a} \in \mathbb{R}^{2k}$ are the learned attention parameters and $f$ is the Leaky Rectified Linear Unit (Leaky ReLU) \cite{leakyrelu} with a negative slope parameter of $\beta~\in~\R^+$.
It is worth noting that the expression $\Matrix{\Theta}\Vector{x}_i$ is actually the $i$-th row of the matrix $\Matrix{X} \Matrix{\Theta}$. 

It is also possible for the GAT operator to have multiple \emph{attention heads}. 
Let us denote the number of heads by $h$. 
Using $h$ attention heads is mathematically equivalent to concatenating results of $h$ GAT operators with different learned parameters. 

Thus, the learned parameters of the layers are:
\begin{equation}
    \mathbf{\Theta} = \Big{[} \mathbf{\Theta}_1 \; ... \; \mathbf{\Theta}_h \Big{]},
\end{equation}

\begin{equation}
    \mathbf{b} = \Big{[} \mathbf{b}_1^T \; ... \; \mathbf{b}_h^T \Big{]^T},
\end{equation}

where $\mathbf{\Theta}_q \in \mathcal{R}^{m \times k}$ denotes the weight matrix for the $q$-th head and $\mathbf{b}_q \in \mathcal{R}^{k}$ represents the bias for $q$-th head.

The formulation for the output feature matrix of the $q$-th head is as follows:

\begin{equation}
        \mathbf{X'}_q = \mathcal{A}_q \mathbf{X} \mathbf{\Theta}_q + \mathbb{1}_{n}\mathbf{b}_q^T,
\end{equation}

where $q$ indicates the head index, $\mathbf{X'}_q \in \mathcal{R}^{n \times k}$ is output feature matrix of the $q$-th head. The matrix $\mathcal{A}_q = (\alpha_{qij}) \in \mathcal{R}^{n \times n}$ represents attention weights of the $q$-th head which are computed as follows:

\begin{equation}
    \alpha^q_{i,j} =
            \frac{
                \exp\left(f\left(\mathbf{a}^T_q
                [\mathbf{\Theta}^T_q\mathbf{x}_i \, \Vert \, \mathbf{\Theta}^T_q\mathbf{x}_j]
                \right)\right)
            }{
                \sum_{k \in \mathcal{N}(i) \cup \{ i \}}
                \exp\left(f\left(\mathbf{a}^{\top}_q
                [\mathbf{\Theta}^T_q\mathbf{x}_i \, \Vert \, \mathbf{\Theta}^T_q\mathbf{x}_k]
                \right)\right)
            },
\end{equation}
where $\mathbf{a}_q \in \mathbb{R}^{2k}$ are the learned attention parameters for the $q$-th head. 

In the end, outputs for all heads are concatenated to create a single output matrix $\Matrix{X'} \in \mathbb{R}^{n \times hk}$:
\begin{equation}
    \mathbf{X'} = \Big{[} \Matrix{X'}_1 \; ... \; \Mat{X'}_h \Big{]}.
\end{equation}

\subsection{GCN Backward Pass}
\label{sec:gcn-bwd}

In order to compute weight updates, we need to compute the gradients of the loss function $\mathcal{L}$ with respect to the parameters of the GCN layer, namely~$\mathbf{\Theta}$ and~$\mathbf{b}$, as well as the gradients with respect to the input node features $\mathbf{X}$, in order to propagate the gradients to the previous layers.

In the derivations, we use the following property. Given some matrices $\Mat{A} \in \R^{a \times b}, \Mat{B} \in \R^{b \times c}$ and operation $\Mat{C} = \Mat{A} \cdot \Mat{B}$, and the gradient of some loss function $\GradL{\Mat{C}}$, we can compute the following gradients:
$$\GradL{\Mat{A}} = \GradL{\Mat{C}} \Mat{B}^T$$
$$\GradL{\Mat{B}} = \Mat{A}^T \GradL{\Mat{C}}$$

Let $\GradL{\mathbf{X'}} \in \mathbb{R}^{n \times k}$ denote the gradient of the loss function with respect to the layer output~$\mathbf{X'}$.
Then we can express the gradient with respect to the parameters $\mathbf{\Theta}$ and $\mathbf{b}$ as follows:

\begin{equation} \label{eq:gcn-bwd-weights}
    \GradL{\mathbf{\Theta}} = \mathbf{X}^T \mathbf{A'}^T\GradL{\mathbf{X'}},
\end{equation}

\begin{equation} \label{eq:gcn-bwd-bias}
    \GradL{\mathbf{b}} = \GradL{\mathbf{X'}}^T  \mathbb{1}_n.
\end{equation}

In case of all layers except the first one, the gradients need to be propagated backward through the model. Thus, we need to also compute the gradient of the loss function with respect to the input node features $\mathbf{X}$. The gradient $\GradL{\mathbf{X}} \in \mathbb{R}^{n \times m}$ can be computed as:

\begin{equation} \label{eq:gcn-bwd-x}
\GradL{\mathbf{X}} = \mathbf{A'}^T  \GradL{\mathbf{X'}} \mathbf{\Theta}^T.
\end{equation}

It is worth noting that matrix product $\Mat{A'}^T \cdot \GradL{\Mat{X'}}$ can be interpreted as propagating the gradients backward through the graph, hence the adjacency matrix is transposed in the backward pass.

Once we have computed the gradients $\GradL{\mathbf{\Theta}}$, $\GradL{\mathbf{b}}$, and $\GradL{\mathbf{X}}$, we can use them to perform a parameter update step using a selected optimizer, such as Stochastic Gradient Descent \cite{sgd} or Adam \cite{adam}.

\subsubsection{GAT Backward Pass}
\label{sec:gat-bwd}

In order to perform a gradient update step, we need to compute gradients of the loss function w. r. t. learned parameters $\Matrix{\Theta}$, $\Vector{a}, \Vector{b}$ and the input features~$\Matrix{X}$, 
respectively denoted by $\GradL{\Matrix{\Theta}}$, $\GradL{\Vector{a}}$, $\GradL{\Vector{b}}$ and $\GradL{\Matrix{X}}$. 

Let us denote $\Matrix{X} \Matrix{\Theta} = \Matrix{M} \in \R^{n \times k}$, the $i$-th row of $\Matrix{M}$ as $\Vector{m}_i$, and $\Vector{a} = \begin{bmatrix}
\Vector{a}_{\text{src}} \\
\Vector{a}_{\text{dst}}
\end{bmatrix}$, where $\Vector{a}_{\text{src}}$, $\Vector{a}_{\text{dst}} \in \R^k$. 
For clarity, let us denote the intermediate values 
$w_{ij}$, 
$y_{ij}$, 
$s_{i} = \Vector{a}_{\text{src}}\Vector{m}_i^T$, and 
$d_{j} = \Vector{a}_{\text{dst}} m_j^T$ 
in the computation of attention weights:
\begin{equation}
    f(\Vector{a}^{\top} [\Matrix{\Theta}\Vector{x}_i \, \Vert \, \Matrix{\Theta}\Vector{x}_j] ) = 
    f(\Vector{a}_{\text{src}}\Vector{m}_i^T + \Vector{a}_{\text{dst}}\Vector{m}_j^T)
    = f(s_{i} + d_{j})
    = f(y_{ij})
    = w_{ij}
\end{equation}

The gradient for $\Vector{b}$ is computed in the same way as in case of GCN:

\begin{equation}\label{eq:gat-bwd-b}  
    \GradL{\Vector{b}} = \GradL{\mathbf{X'}}^T \mathbb{1}_n.
\end{equation}

Then, using $\Matrix{M} = \Matrix{X} \Matrix{\Theta}$, the gradients for $\Mat{X}$ and $\Mat{\Theta}$ can be computed as follows:

\begin{equation}\label{eq:gat-bwd-weights}
    \GradL{\Matrix{\Theta}} = \Matrix{X}^T \GradL{\Matrix{M}},
\end{equation}

\begin{equation}\label{eq:gat-bwd-features}
    \GradL{\Mat{X}} = \GradL{\Mat{M}} \Mat{\Theta},
\end{equation}

where $\GradL{\Mat{M}}$ is the gradient w. r. t. $\Mat{M}$. 
In order to compute $\GradL{\Mat{M}}$, 
$\GradL{\Vector{a_\text{src}}}$, $\GradL{\Vector{a_\text{dst}}}$ we need to compute multiple intermediate gradients.

Firstly, the gradient w. r. t. the attention weights, given that $\Mat{X'} = \mathcal{A} \Mat{M}$:
\begin{equation}
    \GradL{\mathcal{A}} = \GradL{\Mat{X'}} \Mat{M}^T
\end{equation}

Then, we need to backpropagate through the column-wise softmax. 
\begin{equation}
    \GradL{w_{ij}} = \alpha_{ij} \Big(\GradL{\alpha_{ij}} - \sum_{u = 1}^{n} \alpha_{iu} \GradL{\alpha_{iu}} \Big)
\end{equation}

The next step is computing the derivative of Leaky ReLU. Its derivative is: 
\begin{equation}
f'(x) = \begin{cases}
1 & \text{if } x > 0, \\
\beta & \text{if } x \leq 0.
\end{cases}
\end{equation}

Then the gradient is as follows from the chain rule.
\begin{equation}
    \GradL{y_{ij}} = f'(y_{ij}) \cdot \GradL{w_{ij}}
\end{equation}

Now we just need to compute the gradients for $\Vector{s} = \Vector{a}_{\text{src}}\Mat{M}^T$ and $\Vector{d} = \Vector{a}_{\text{dst}}\Mat{M}^T$. Given that $y_{ij} = s_i + d_j$, computing the gradients just requires a summation along an appropriate axis.

\begin{equation}
    \GradL{s_{i}} = \sum_{u=1}^n \GradL{y_{iu}},
    \quad\quad
    \GradL{d_j} = \sum_{u=1}^n \GradL{y_{uj}}
\end{equation}

Now we can use $\GradL{\Vector{d}}$, $\GradL{\Vector{s}} \in \R^n$ in order to compute the gradients of the attention parameters:

\begin{equation}
    \GradL{\Vector{a_\text{src}}} = \Mat{M}^T \GradL{\Vector{s}}, 
    \quad\quad 
    \GradL{\Vector{a_\text{dst}}} = \Mat{M}^T \GradL{\Vector{d}}.
\end{equation}

Finally, having $\GradL{\Vector{d}}$, $\GradL{\Vector{s}}$  also allows us to compute the gradient for $\Mat{M}$.

\begin{equation}
    \GradL{\Mat{M}} = 
    \Big(\GradL{\Vector{d}}\Big)^T \cdot \Vector{a_\text{dst}} 
    + \Big(\GradL{\Vector{s}}\Big)^T \cdot \Vector{a_\text{src}} 
    + \mathcal{A}^T \GradL{\Mat{X}'}
\end{equation}

Plugging this value into Equations \ref{eq:gat-bwd-weights} and \ref{eq:gat-bwd-features}, we are able to obtain the gradients for the remaining parameters.

\section{Figures}

\begin{figure}[hbtp]
    \centering

    \includegraphics[width=\textwidth]{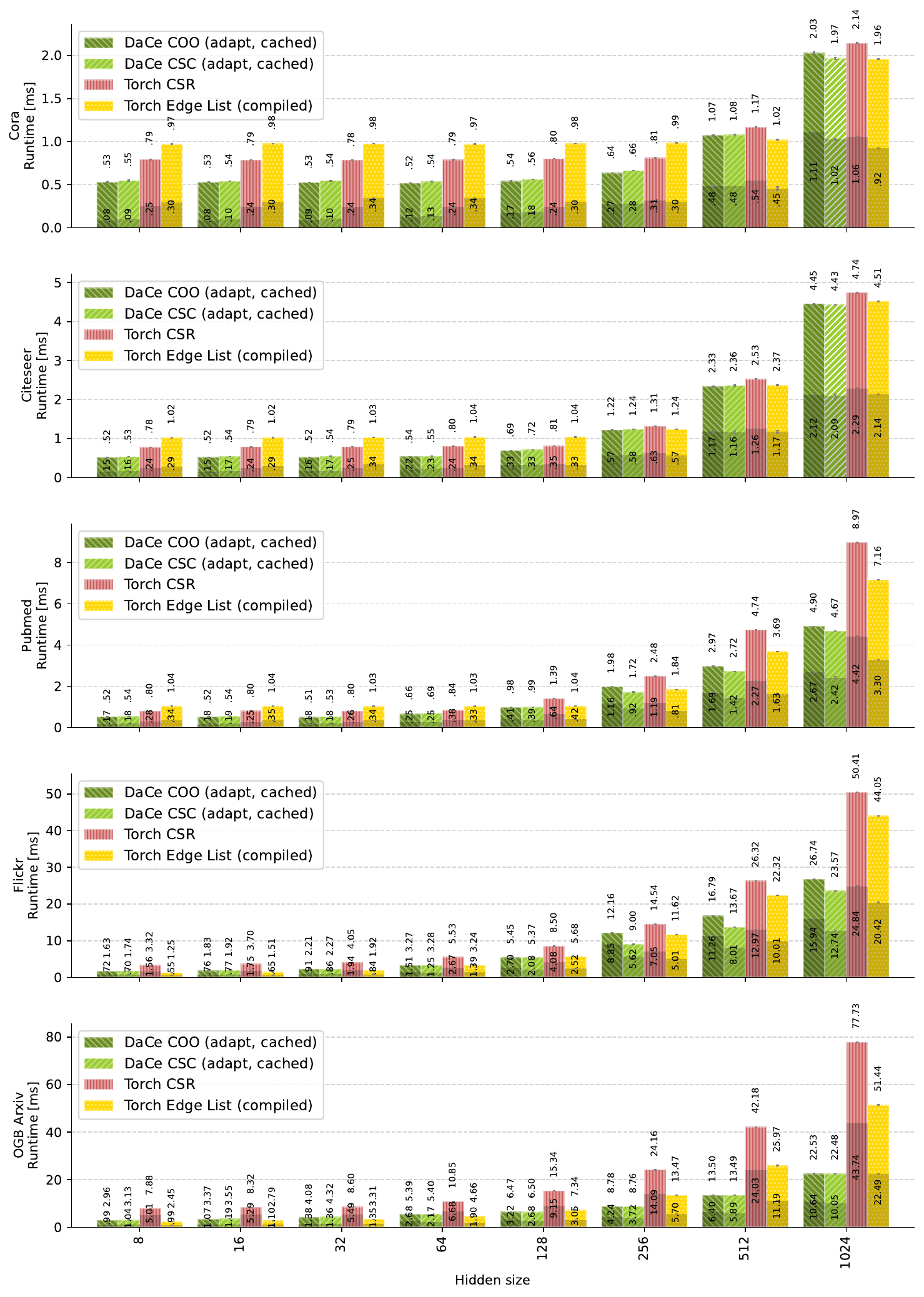}

    \caption{Detailed GCN runtime results, comparison of PyTorch Geometric and our work. Darker regions indicate the forward pass, total bar height is the sum of forward, loss and backward runtime.}
    \label{fig:gcn-main-results}
\end{figure}

\begin{figure}
    \centering
    \includegraphics[width=0.8\textwidth]{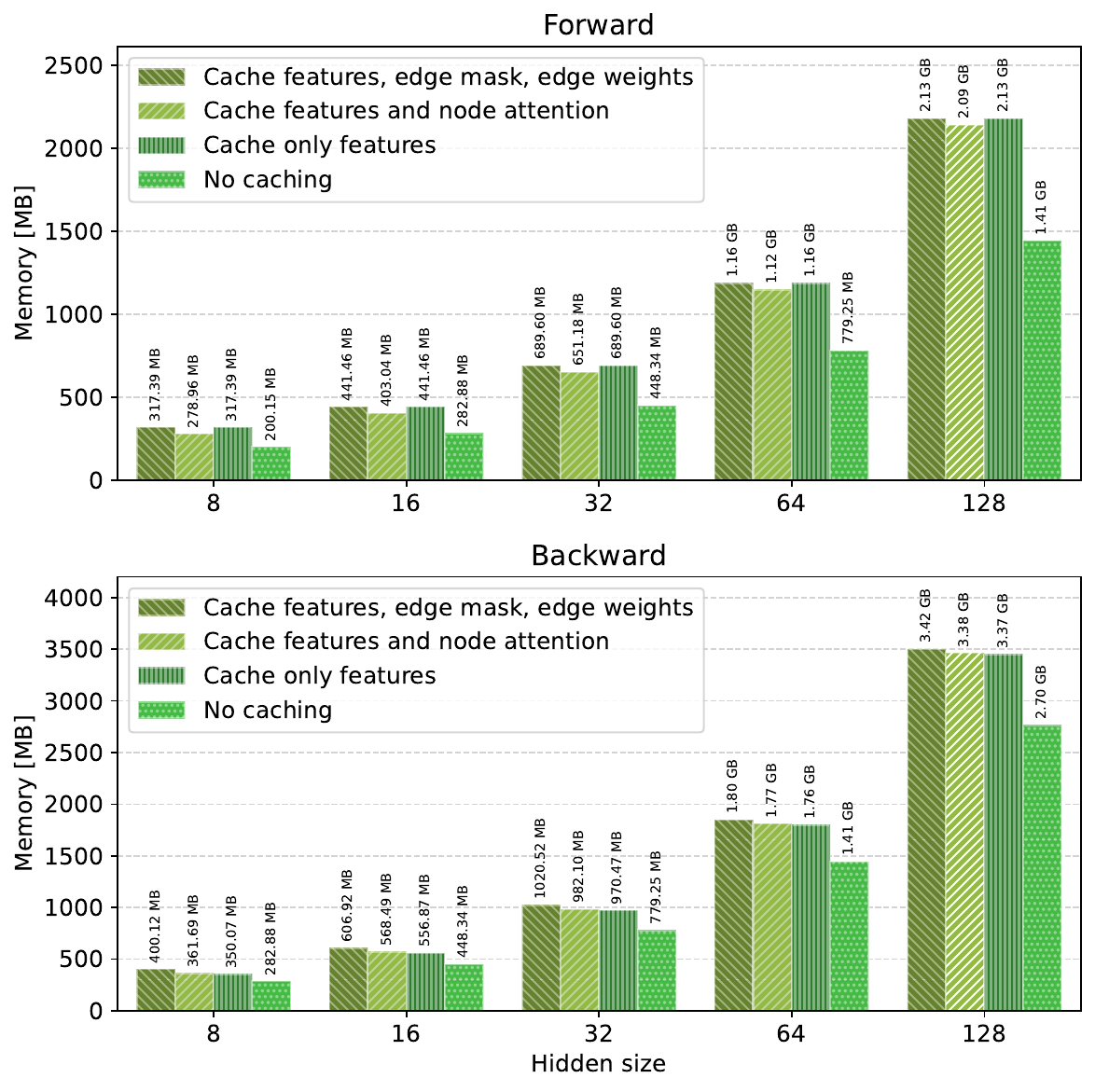}
    \caption{GAT memory use of different caching schemes on the OGB Arxiv dataset.}
    \label{fig:exp-gat-mem}
\end{figure}

\begin{figure}[hbtp]
    \centering
    \begin{subfigure}{0.49\textwidth}
        \centering
        \includegraphics[width=\textwidth]{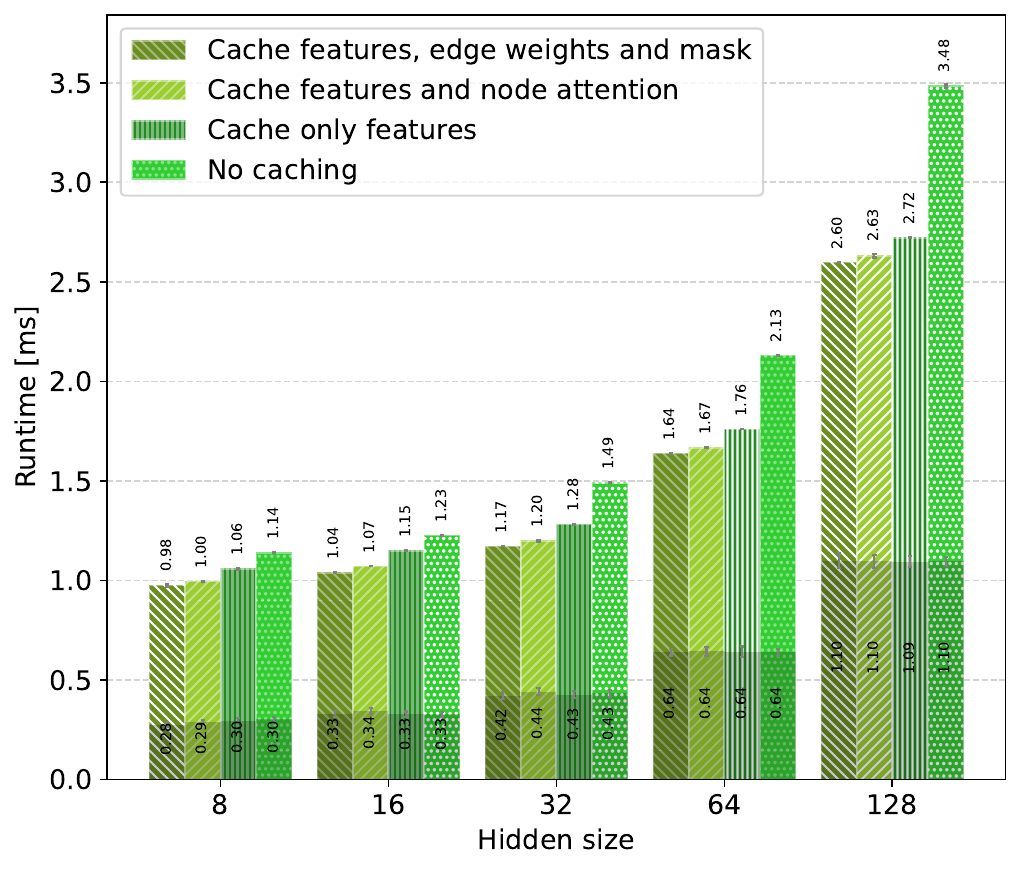}
        \caption{GAT runtime, Cora dataset.}
        \label{subfig:gat-cora}
    \end{subfigure}
    \begin{subfigure}{0.49\textwidth}
        \centering
        \includegraphics[width=\textwidth]{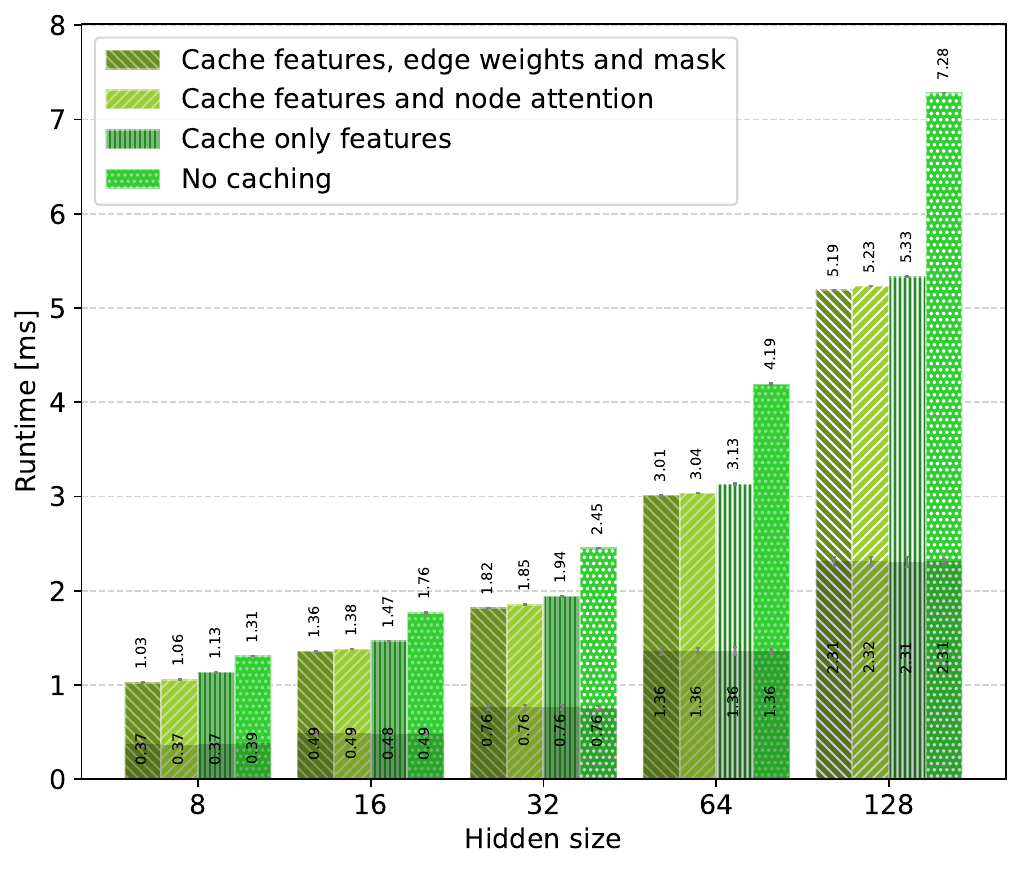}
        \caption{GAT runtime, Citeseer dataset.}
        \label{subfig:gat-citeseer}
    \end{subfigure}
    \begin{subfigure}{0.49\textwidth}
        \centering
        \includegraphics[width=\textwidth]{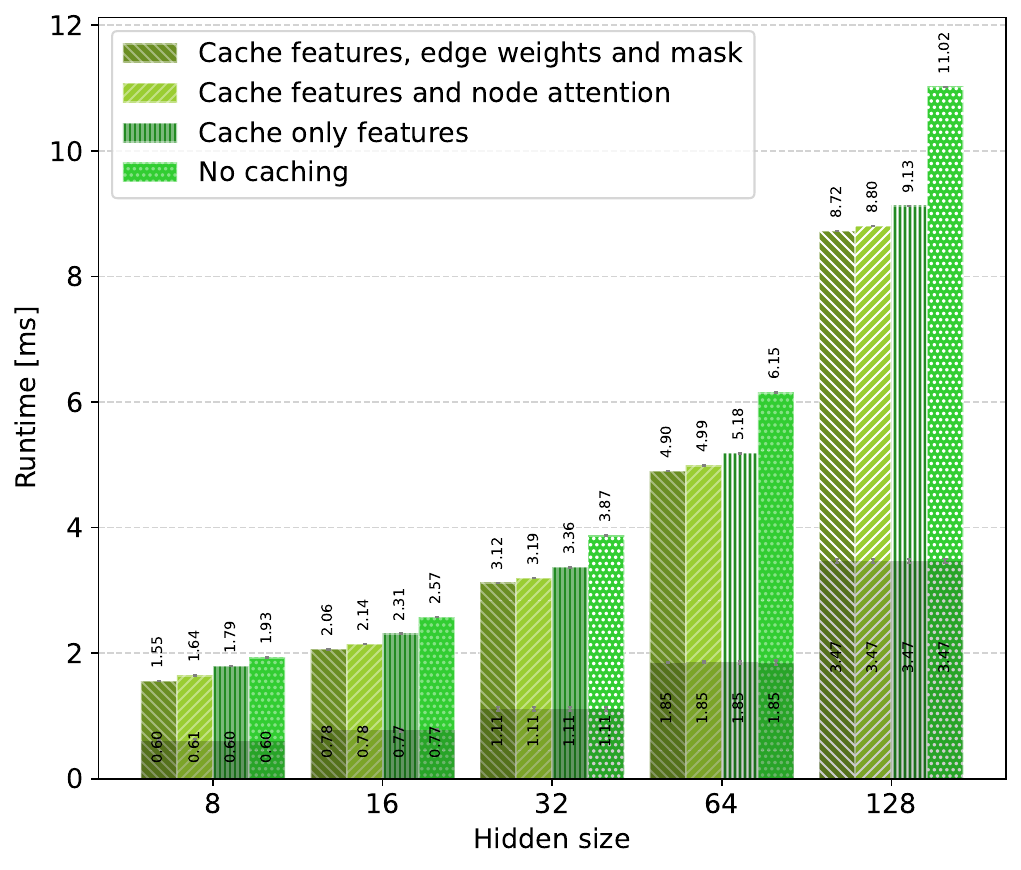}
        \caption{GAT runtime, Pubmed dataset.}
        \label{subfig:gat-pubmed}
    \end{subfigure}
    \begin{subfigure}{0.49\textwidth}
        \centering
        \includegraphics[width=\textwidth]{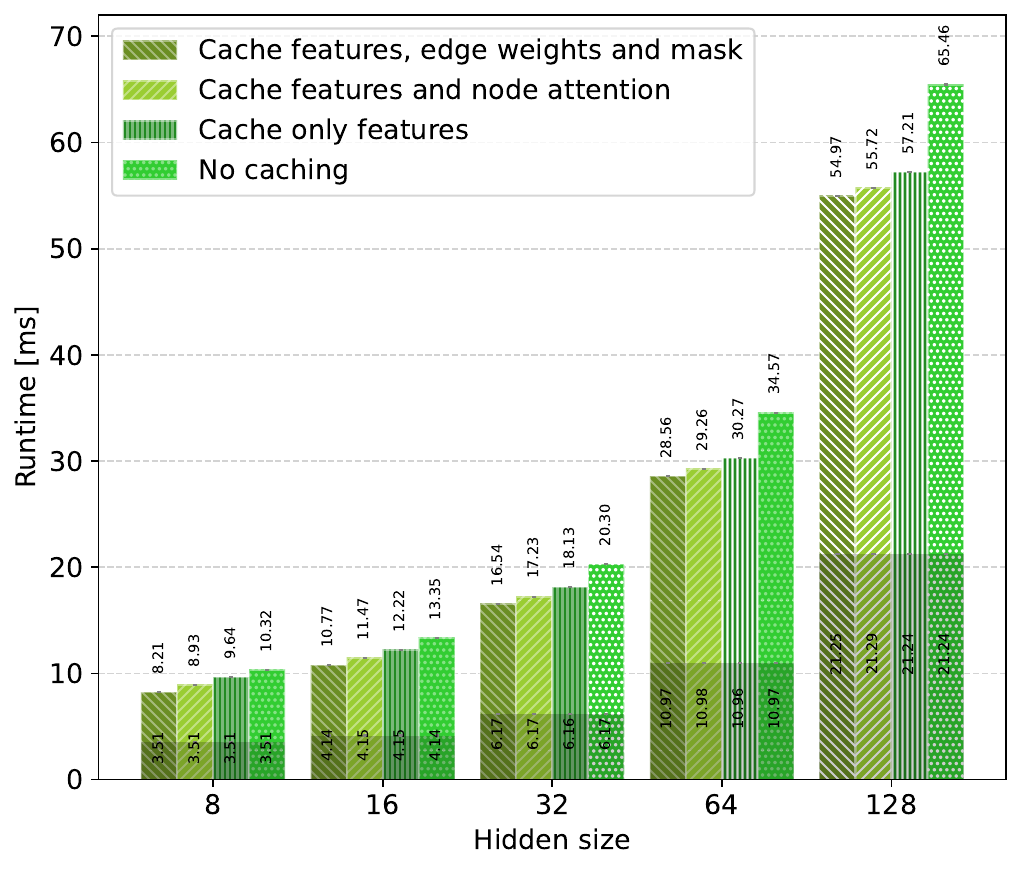}
        \caption{GAT runtime, Flickr dataset.}
        \label{subfig:gat-flickr}
    \end{subfigure}
    \begin{subfigure}{0.5\textwidth}
        \centering
        \includegraphics[width=1\textwidth]{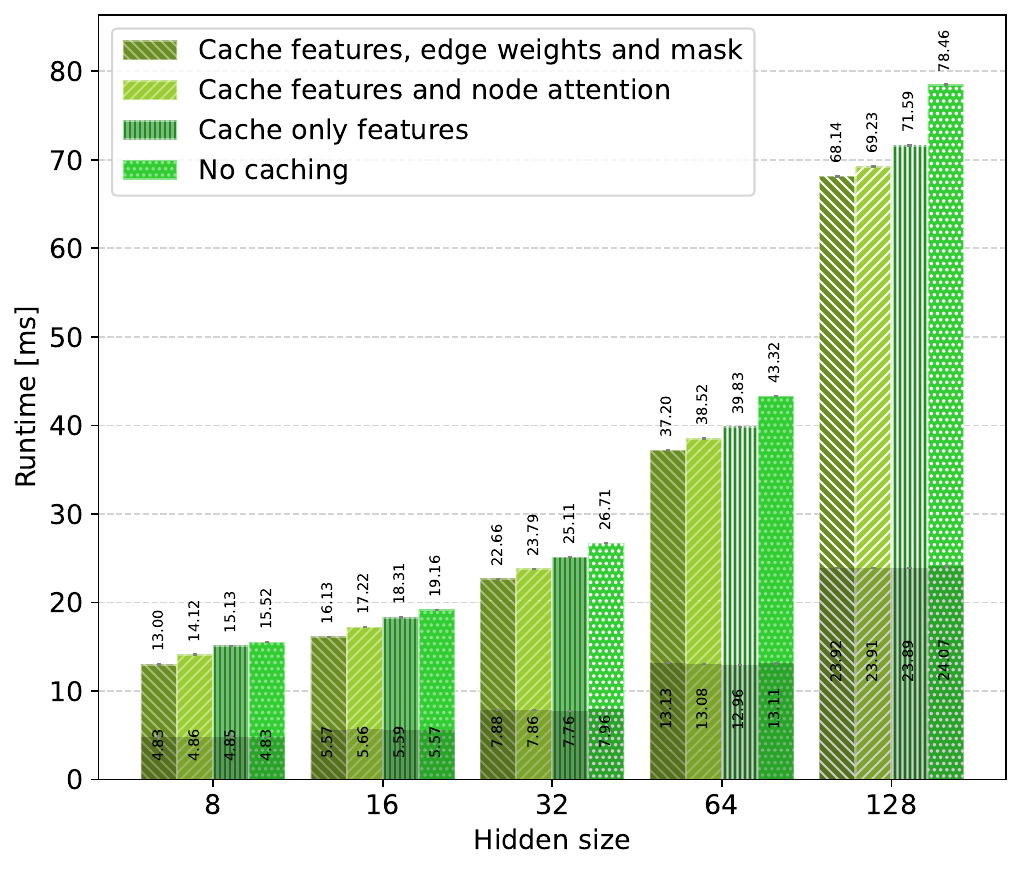}
        \caption{GAT runtime, OGB Arxiv dataset.}
        \label{subfig:gat-arxiv}
    \end{subfigure}
    \caption{Detailed GAT runtime results, comparison of various caching schemes. Darker regions indicate the forward pass, total bar height is the sum of forward, loss and backward runtime.}
    \label{fig:exp-gat-caching-results}
\end{figure}

\begin{figure}[hbtp]
    \centering
    \begin{subfigure}{0.48\textwidth}
        \centering
        \includegraphics[width=\textwidth]{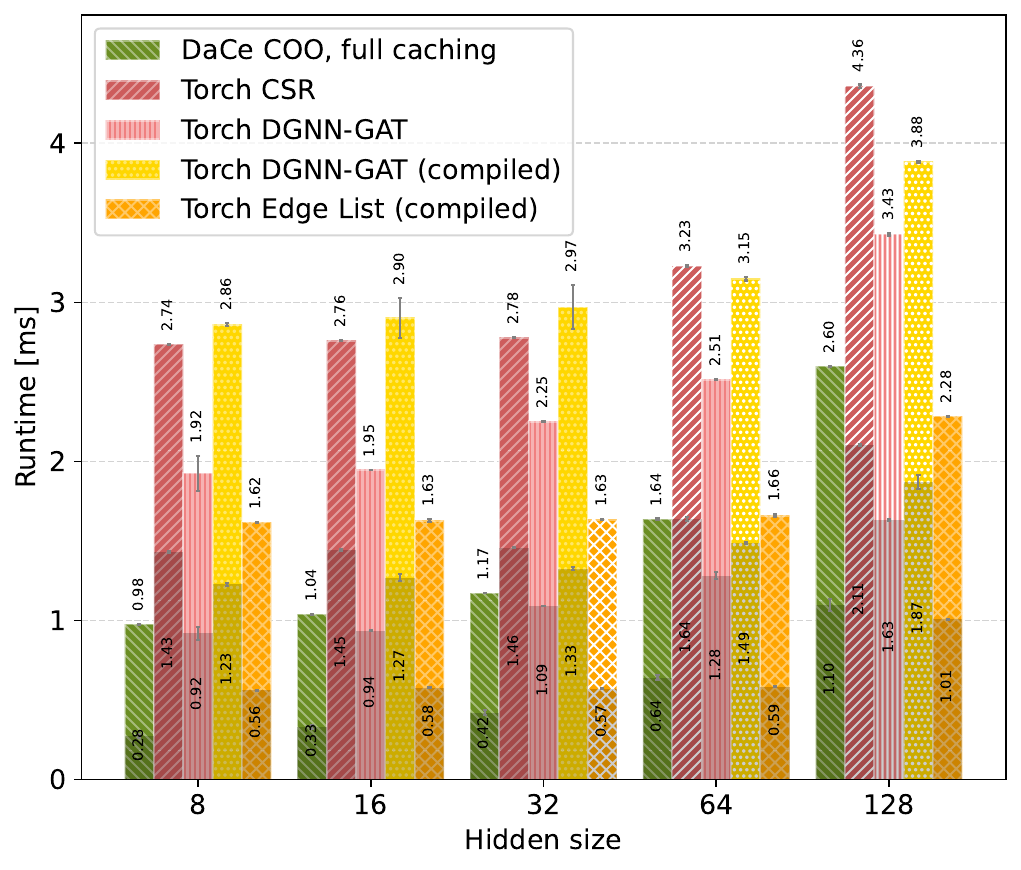}
        \caption{GAT runtime, Cora dataset.}
    \end{subfigure}
    \begin{subfigure}{0.48\textwidth}
        \centering
        \includegraphics[width=\textwidth]{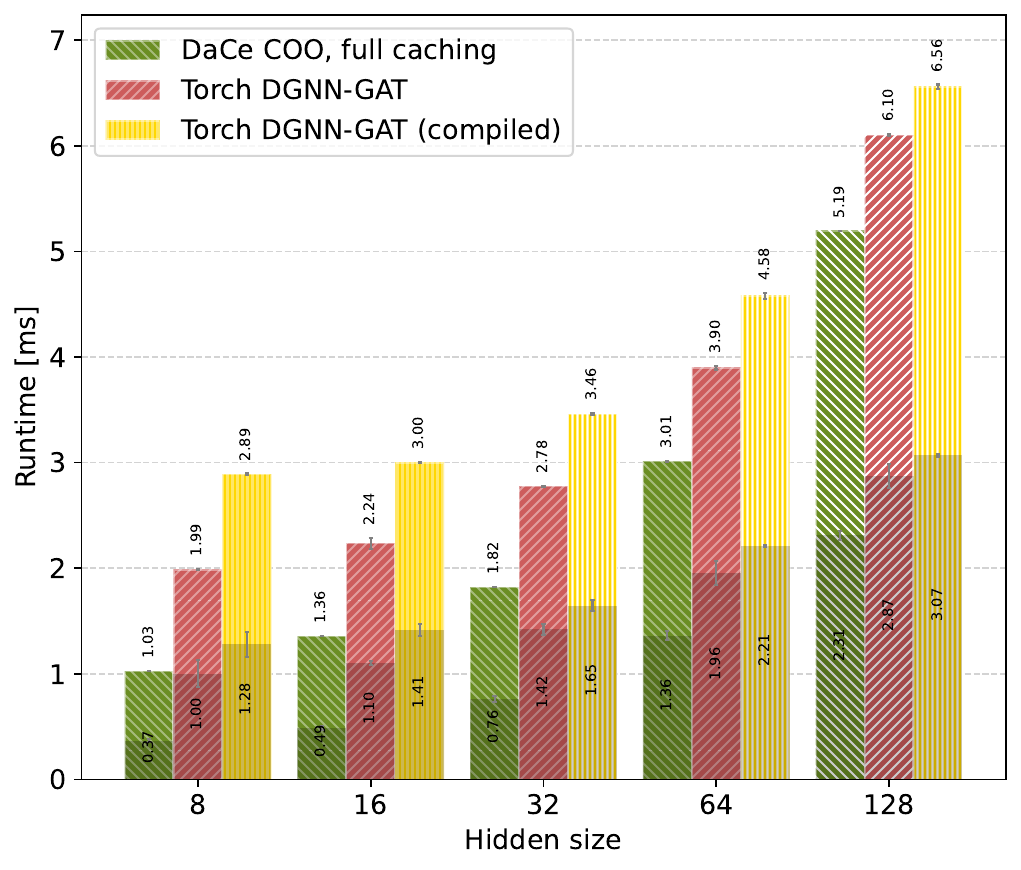}
        \caption{GAT runtime, Citeseer dataset.}
    \end{subfigure}
    \begin{subfigure}{0.48\textwidth}
        \centering
        \includegraphics[width=\textwidth]{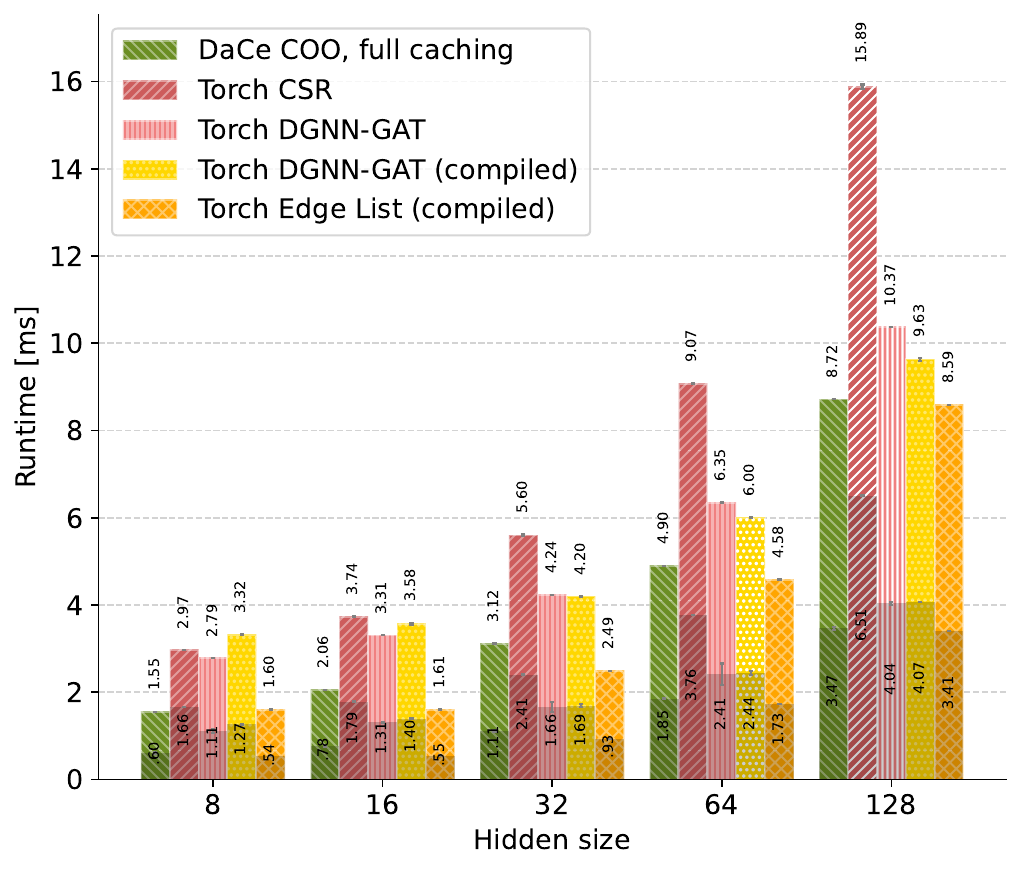}
        \caption{GAT runtime, Pubmed dataset.}
    \end{subfigure}
    \begin{subfigure}{0.48\textwidth}
        \centering
        \includegraphics[width=\textwidth]{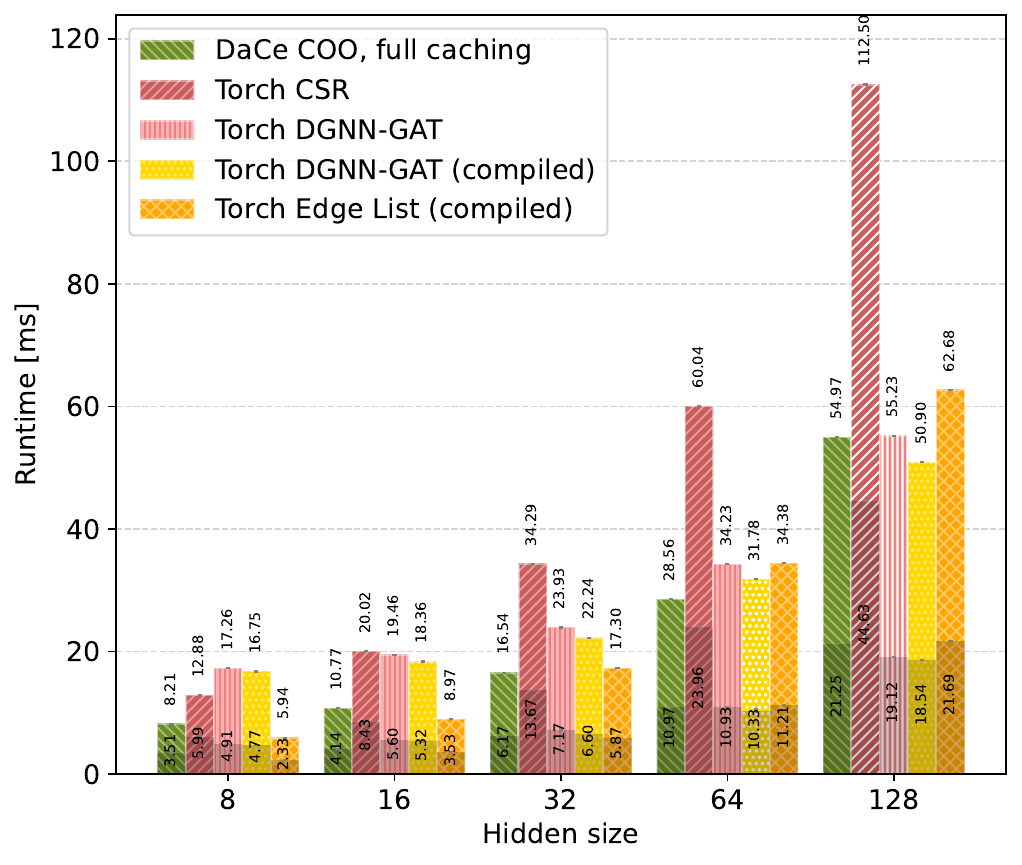}
        \caption{GAT runtime, Flickr dataset.}
    \end{subfigure}
    \begin{subfigure}{0.48\textwidth}
        \centering
        \includegraphics[width=\textwidth]{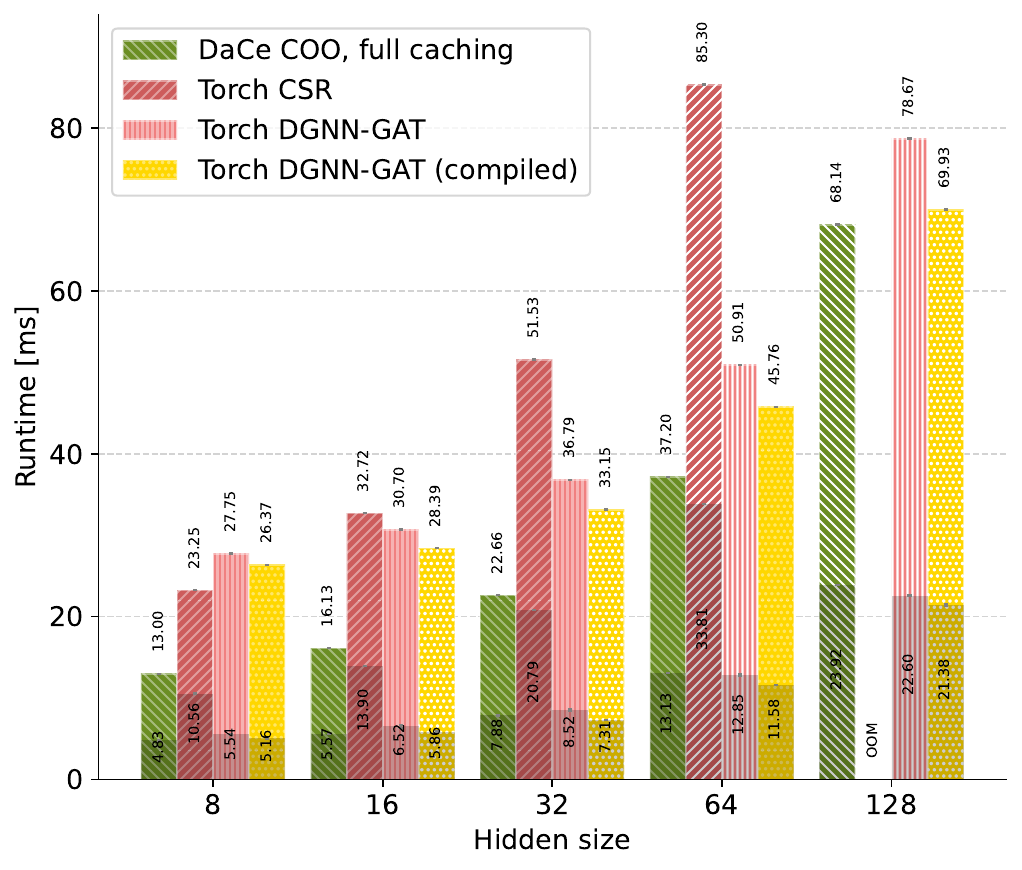}
        \caption{GAT runtime, OGB Arxiv dataset.}
    \end{subfigure}
    \caption{Detailed GAT runtime results, comparison of our DaCe implementation against PyTorch. Darker regions indicate the forward pass, total bar height is the sum of forward, loss and backward runtime. Some data points for the compiled PyTorch implementation using edge lists are missing due to compilation errors.}
    \label{fig:exp-gat-baselines-results}
\end{figure}

\end{document}